\documentclass[10pt,journal,compsoc]{IEEEtran}
\usepackage[numbers,sort&compress]{natbib}

\usepackage{tablefootnote}
\usepackage{graphicx}
\usepackage{amsmath}
\usepackage{amssymb}
\usepackage{booktabs}
\usepackage{float}
\usepackage{multirow}
\usepackage{array}
\usepackage{mathrsfs}
\usepackage{subfigure}
\usepackage{xcolor}
\usepackage{soulpos}
\usepackage{url}
\usepackage{array}
\usepackage{booktabs}
\usepackage{amsthm}
\usepackage{arydshln}
\usepackage{multirow}
\usepackage{tabularx}
\usepackage{soul}
\usepackage{makecell}
\usepackage{contour}
\usepackage{caption}
\usepackage{enumitem}
\usepackage{subcaption}

\ulposdef{\hlc}[xoffset=1pt]{\mbox{\color{orange!30}\rule[-.8ex]{\ulwidth}{3ex}}}
\ulposdef{\be}[xoffset=1pt]{\mbox{\color{orange!80}\rule[-.8ex]{\ulwidth}{2.4ex}}}
\ulposdef{\bee}[xoffset=1pt]{\mbox{\color{orange!60}\rule[-.8ex]{\ulwidth}{2.4ex}}}
\ulposdef{\beee}[xoffset=1pt]{\mbox{\color{orange!40}\rule[-.8ex]{\ulwidth}{2.4ex}}}
\ulposdef{\beeee}[xoffset=1pt]{\mbox{\color{orange!20}\rule[-.8ex]{\ulwidth}{2.4ex}}}
\ulposdef{\beeeee}[xoffset=1pt]{\mbox{\color{orange!5}\rule[-.8ex]{\ulwidth}{2.4ex}}}

\ifCLASSOPTIONcompsoc
\else
\usepackage{cite}
\fi
\ifCLASSINFOpdf
\else
\fi
\begin{document}
\title{A Survey of Stance Detection on Social Media: New Directions and Perspectives}
\author{Bowen~Zhang$^\dag$,
Genan~Dai$^\dag$,
Fuqiang~Niu,
% Xueqi~Cheng,
% Xianghua~Fu,
Nan~Yin,
Xiaomao~Fan,
Senzhang Wang,\\
Xiaochun Cao, \textit{Senior Member}, \textit{IEEE},
and~Hu~Huang$^*$
% 	%        John~Doe,~\IEEEmembership{Fellow,~OSA,}
% 	%        and~Jane~Doe,~\IEEEmembership{Life~Fellow,~IEEE}% <-this % stops a space
% 	% Leung, Ka Cheong
\thanks{Bowen Zhang, Genan Dai and Xiaomao Fan are with the College of Big Data and Internet, Shenzhen Technology University, Shenzhen, 518000, China.}
\thanks{Fuqiang Niu is with the College of Computer Science and Software, Shenzhen University, Shenzhen, 518000, China.}
\thanks{Nan Yin is affiliated with Hong Kong University of Science and Technology, Hong Kong, 999077, China.}
\thanks{Senzhang Wang is with the School of Computer Science and Engineering, Central South University, Changsha 410083, China.}
\thanks{Xiaochun Cao is with the School of Cyber Science and Technology, Shenzhen Campus of Sun Yat-sen University, Shenzhen 518107, China.}
\thanks{Hu Huang is affiliated with the Peking University Shenzhen Graduate School, Shenzhen, 518000, China.}
\thanks{$^{\dag}$ Equal contribution.}
\thanks{$^*$ Corresponding author: Hu Huang (Email: huanghu@mail.ustc.edu.cn).}
\thanks{Manuscript received NOV 25, 2024.}
}


\markboth{Journal of \LaTeX\ Class Files,~Vol.~14, No.~8, August~2015}%
{Shell \MakeLowercase{\textit{et al.}}: Bare Demo of IEEEtran.cls for Computer Society Journals}



\IEEEtitleabstractindextext{%
\begin{abstract}
In modern digital environments, users frequently express opinions on contentious topics, providing a wealth of information on prevailing attitudes. The systematic analysis of these opinions offers valuable insights for decision-making in various sectors, including marketing and politics. As a result, stance detection has emerged as a crucial subfield within affective computing, enabling the automatic detection of user stances in social media conversations and providing a nuanced understanding of public sentiment on complex issues.
Recent years have seen a surge of research interest in developing effective stance detection methods, with contributions from multiple communities, including natural language processing, web science, and social computing. 
This paper provides a comprehensive survey of stance detection techniques on social media, covering task definitions, datasets, approaches, and future works. We review traditional stance detection models, as well as state-of-the-art methods based on large language models, and discuss their strengths and limitations.
Our survey highlights the importance of stance detection in understanding public opinion and sentiment, and identifies gaps in current research. We conclude by outlining potential future directions for stance detection on social media, including the need for more robust and generalizable models, and the importance of addressing emerging challenges such as multi-modal stance detection and stance detection in low-resource languages. 
\end{abstract}

}

\maketitle

\IEEEdisplaynontitleabstractindextext

% \fi
\IEEEpeerreviewmaketitle

\IEEEraisesectionheading{\section{Introduction}\label{sec:introduction}}

% Social media platforms have increasingly served as a common forum for users to express their viewpoints on controversial issues. By aggregating and analyzing these perspectives, we can identify prevailing trends and opinions on contentious topics, ranging from abortion to epidemic prevention. This wealth of data offers significant potential for web mining and content analysis. The insights gathered from such analysis can be crucial for various decision-making processes, including advertising strategies and political elections. Consequently, automatic stance detection on social media has gained importance in opinion mining, facilitating a more nuanced understanding of user attitudes towards a wide array of issues.
The proliferation of social media platforms has revolutionized modern communication, transforming these platforms into a primary source of news and information for individuals. By facilitating user-generated content, these platforms enable individuals to articulate their opinions, share perspectives, and explore diverse facets of emerging topics~\cite{glandt2021stance,giasemidis2018semi}. 
The widespread adoption of social media has yielded a treasure trove of data for researchers to investigate various aspects of online human behavior, including the public's stance on social and political issues~\cite{li2021p}. Through the analysis and aggregation of user perspectives, researchers can identify dominant trends and opinions on polarizing topics. This vast repository of data holds significant potential for web mining and content analysis, with the insights gleaned from such analysis having far-reaching implications for decision-making processes, including advertising strategies and political elections~\cite{niu2024challenge,MohammadKSZC16}. Consequently, automated stance detection on social media has become a crucial aspect of opinion mining, enabling a more nuanced understanding of user attitudes towards a diverse range of issues \cite{chen2018we}.

The aim of stance detection is to determine the stance of a text's author towards a target, which can be an entity, concept, event, idea, opinion, claim, or topic that is either explicitly mentioned or implied within the text~\cite{mohammad2017stance,sobhani-etal-2017-dataset}.
Early research on stance detection primarily focused on analyzing debates on online forums, which differs from more recent work that has emphasized social media platforms~\cite{pontiki2016semeval}, particularly $X$
(Twitter)\footnote{https://x.com/home}. Online forum debates typically have a clear, single context compared to social media platforms. In forum discussions, users engage in threaded conversations where information flow is usually focused on a specific topic. In contrast, social media discussions on a given topic are more dispersed, although they can sometimes be linked through hashtags\footnote{Hashtags are words and numbers following the \# symbol that categorize and track content on social media.}. However, social media platforms offer a wealth of additional features, such as user networks, which web science and computational social science researchers have utilized to develop stance detection techniques that do not rely solely on text~\cite{addawood2017stance,küçük2017stance}.

\textbf{Related surveys.} 
Recently, several research surveys on stance detection have been conducted. \citet{ZubiagaABLP18} focused on stance detection in product reviews, with a particular emphasis on feature-level stance detection techniques. \citet{kuccuk2020stance} provided a comprehensive review of stance detection definitions, task classifications, and the relationships between stance detection and related subtasks, primarily focusing on natural language processing (NLP) based methods. \citet{aldayel2021stance} were the first to survey stance detection techniques for social media, covering not only NLP techniques but also network-level and multi-modal information, providing a more comprehensive overview of stance detection research in social media.

\textbf{Our motivation.} In this paper, we present an overview of stance detection on social media and discuss promising research directions. The key motivations of this survey are summarized as follows:

\begin{itemize}
\item 
Recent advances in stance detection have shown significant progress. However, existing surveys primarily concentrate on conventional approaches, including feature engineering, traditional neural networks, and early pre-trained language models such as BERT. These reviews have not comprehensively analyzed the potential of emerging large language models (LLM), particularly the GPT family, in stance detection tasks.
Therefore, to comprehensively survey and synthesize the state-of-the-art, it is imperative to consolidate and distill these recent works.

\item 
The rapid evolution of social media has led to an increasing presence of multi-modal information and interactive patterns, rendering traditional text-based stance detection methods inadequate for these emerging contexts. In response, recent years have witnessed a proliferation of novel tasks specifically designed for social media, underscoring the need for a comprehensive review and synthesis of these advances.

\item 
Despite the promising potential of LLMs for stance detection, their application is still in an initial exploratory phase. Effective techniques to adapt and utilize these models for specific stance detection tasks remain an open research challenge. Moreover, the roles, background knowledge understanding, and biases of LLMs in stance detection contexts need to be further examined. Consequently, consolidating findings on these outstanding issues will provide significant motivation for authoring a new survey paper.

\item As LLM continues to advance, stance detection applications are expanding. While traditionally focused on sentence-level textual analysis, stance detection now shows potential for richer data types and settings. Moreover, new domains beyond text could benefit from stance detection. 
Looking ahead, it is necessary to discuss potential research directions that can improve stance detection and mitigation capabilities.
\end{itemize}

\textbf{Methodology for literature review.}
For our literature review, we conducted a search on Google Scholar, DBLP, and Web of Science using the keywords ``stance detection/prediction/classification'' and other related terms such as ``viewpoint'' and ``perspective''. We focused on studies where the search keywords were mentioned in the title or abstract. The search results were refined using various parameters, including the number of citations, year of publication, and the ranking of the journal or conference. We placed a special emphasis on venues related to NLP, computational social science, Web science, and information processing. The list of relevant studies was then inspected, and additional relevant studies were identified from the reference lists of those papers and the papers that cited them. We also conducted an extensive inspection of papers citing the SemEval 2016~\cite{MohammadKSZC16} stance detection task.

% One of the main observations from a review of the stance detection literature is the significant increase in the number of studies on this topic in recent years. Fig. 1 illustrates the number of publications on stance detection from 2008 to 2020, as indexed by Web of Science. The figure shows a steady increase in the number of publications, with a sudden increase from 2016. This increase is attributed to the release of the SemEval 2016 stance detection task \cite{MohammadKSZC16}, which provided the first benchmark dataset for stance detection on social media, particularly for the NLP community. This dataset has facilitated research in stance detection on social media and opened up opportunities for the development of methodologies for stance representations on social media.

The remainder of this survey is structured as follows. 
Section 2 provides an overview of related tasks in stance detection. Section 3 presents datasets for stance detection under various tasks. 
We then review traditional stance detection methods in Section 4, tracing the technical development trajectory. 
Next, we introduce state-of-the-art stance detection methods based on LLMs in Section 5. 
Finally, we discuss the challenges faced by stance detection in the era of LLM in Section 6, as well as future research directions and potential development paths.

\section{Stance Detection Tasks}
% According to the literature, stance detection tasks can be categorized from multiple perspectives. 
% From a data-centric view, stance detection can be divided into subtasks based on target type, including specific-target stance detection, multi-target stance detection, and claim-based stance detection. 
% Additionally, from a data content perspective, stance detection can be classified into conversational-based, multi-modal, and user-level stance detection tasks.
% From a model deployment perspective, stance detection tasks can be categorized into in-target, cross-target, zero-shot, and cross-lingual stance detection tasks.
% \section{Stance Detection Tasks}
According to the literature, stance detection tasks can be categorized from two primary perspectives: data-centric and method development.
From a data-centric perspective, stance detection encompasses several subtasks: target-specific stance detection, multi-target stance detection, claim-based stance detection, conversational stance detection, multi-modal stance detection, and user-level stance detection.
From a method development perspective, the tasks can be classified into in-target stance detection, cross-target stance detection, zero-shot stance detection, and cross-lingual stance detection, each addressing different deployment scenarios and technical challenges.

\subsection{Data-Centric Perspective}
From the perspectives of target types, data modalities, and categories, this section provides a detailed introduction to related tasks.

\textbf{Target-specific stance detection.}
The target-specific stance detection task is the most common task setting in the literature. In this paradigm, stance is inferred with respect to a predefined set of targets. Most previous studies have focused on this paradigm, including works such as ~\citet{augenstein2016stance}, ~\citet{darwish2017improved}, ~\citet{igarashi-etal-2016-tohoku}, ~\citet{liu2017attention}, and ~\citet{MohammadKSZC16}. In this approach, the input text is used to predict the stance towards a specific, predefined single target.

\begin{table*}
  \caption{Target-specific stance detection  datasets}
  \label{intar-dataset}
  \resizebox{\linewidth}{!}{
  \footnotesize
  \begin{tabular}{>{\raggedright\arraybackslash}p{0.18\linewidth}>{\raggedright\arraybackslash}p{0.25\linewidth}>{\raggedright\arraybackslash}p{0.1\linewidth}>{\raggedright\arraybackslash}p{0.1\linewidth}ll>{\raggedright\arraybackslash}p{0.18\linewidth}}
    % \hline
    \toprule
    \textbf{Dataset-name/Authors}& \textbf{Target(s)} & \textbf{Source} & \textbf{Size} & \textbf{Time} & \textbf{Language} &\textbf{Annotation Classes}\\
    \midrule
%  \citet{thomas-etal-2006-get}& Proposed legislations& Online political
% debates& 3,857 speech segments and 53 debates& 2006& English&Yes, No\\
% \midrule
%  \citet{somasundaran-wiebe-2009-recognizing}& Firefox vs. IE, iPhone vs. Blackberry, Opera vs. Firefox,  Sony
% Ps3 vs. Nintendo Wii, Windows vs.
% Mac& Online debates on products& 304 debate posts& 2009& English&Pro-Firefox...\\
% \midrule

%  \citet{somasundaran2010recognizing}& Several topics in healthcare, Existence of God, Gun rights, Gay rights, Abortion, and Creationism& Online ideological debates & 7,134 debate
% & 2010& English&For, Against\\
% \midrule

%  \citet{murakami-raymond-2010-support}& Five ideas& Online debates& 481 comments& 2010& Japanese&Support, Oppose\\
% \midrule
 SemEval-2016 Task 6~\cite{MohammadKSZC16} & Atheism, Climate Change is Concern, Feminist Movement, Hillary Clinton, Legalization of Abortion, Donald Trump& $X$ (Twitter)  & 4,870 tweets& 2016 & English&Favor, Against, Neither\\
% \midrule
 
  NLPCC-2016~\cite{xu2016overview}& iPhone SE, Set off firecrackers in the Spring Festival, Russia’s anti-terrorist operations in Syria, Two-child policy, Prohibition of motorcycles and restrictions on electric vehicles in Shenzhen, Genetically modified food, Nuclear test in DPRK & Weibo & 4,000 annotated and 2,400 unannotated tweets& 2016& Chinese&Favor, Against, None\\
% \midrule
  
  % IAC 2.0~\cite{abbott-etal-2016-internet}& Various topics& Online debates& 482 posts& 2016&   English&Pro, Con\\
% \midrule
  
\citet{küçük2017stance}& Galatasaray, Fenerbahçe& $X$ (Twitter)  & 700 tweets& 2017& Turkish&Favor, Against\\
% \midrule

    SemEval-2017~\cite{mohammad2017stance}& Atheism, Climate Change is Concern, Feminist Movement, Hillary Clinton, Legalization of Abortion, Donald Trump& $X$ (Twitter)  & 4,870 tweets& 2017&   English&Favor, Against, Neither for stance; Positive, Negative, and Neither for sentiment\\
% \midrule
    
 \citet{addawood2017stance}& Individual privacy, Natural
security, Other, Irrelevant& $X$ (Twitter)  & 3,000 tweets& 2017& English&Favor, Against, Neutral\\
% \midrule

 \citet{darwish2017improved}& Transfer of two islands from Egypt to Saudi Arabia& $X$ (Twitter)  & 33,024 tweets& 2017& Arabic&Favor (Positive), Against (Negative)\\
% \midrule
 
 \citet{hercig2017detecting}& Miloš Zeman, Smoking ban in restaurants& iDNES.cz& 5,423 news comments& 2017& Czech&Support, Deny, Query, Comment\\
% \midrule
 
 % \cite{lai2018stance}& 2016 referendum on reform of the Italian Constitution& $X$ (Twitter)  & 993 triplets (2,889 tweets)& 2018& Italian&Favor, Against, None\\
 \citet{küçük2018stance}& Galatasaray, Fenerbahçe& $X$ (Twitter)  & 1,065 tweets& 2018& Turkish&Favor, Against\\
% \midrule
 Me Too dataset~\cite{gautam2020metooma}& Me Too movement& $X$ (Twitter)  & 9,973 tweets& 2019& English&Supprt, Opposition\\
% \midrule
    TSE2020~\cite{kawintiranon-singh-2021-knowledge} & Donald Trump, Joe Biden& $X$ (Twitter)  & 2,500 tweets& 2020 &   English&Favor, Against, Neither\\
% \midrule
    
    P-stance~\cite{li2021p} & Donald Trump, Joe Biden, Bernie Sanders & $X$ (Twitter) & 21,574 tweets& 2020&   English&Favor, Against, Neither\\
% \midrule
    
    WT-WT~\cite{conforti2020will} & Aetna, Express Scripts, Cigna, Humana, 21st Century Fox & $X$ (Twitter) & 51,284 tweets& 2020& 
      English&Favor, Against, Neither\\
% \midrule

    COVID-19-Stance~\cite{glandt2021stance} & Anthony S. Fauci, M.D., Keeping Schools Closed, Stay At Home Orders, Wearing a Face Mask & $X$ (Twitter) & 7,122 tweets& 2020&  English&Favor, Against, Neither\\
% \midrule

    VAST~\cite{allaway2020zero} & Various topics& The New York Times& 23,525 tweets& 2020&   English&Favor, Against, Neither\\
% \midrule
    Mawqif~\cite{alturayeif-etal-2022-mawqif} & COVID-19 vaccine, digital transformation, women empowerment& $X$ (Twitter)&4,121 tweets& 2022&   Arabic &Favor, Against, None\\
% \midrule
    C-STANCE Subtask A~\cite{zhao-etal-2023-c}& Covid Epidemic, World Events, Cultural and Education, Entertainment and Consumption, Sports, Rights, Environmental Protection& Weibo &19,038 tweets & 2023&   Chinese &Favor, Against, Neutral\\
% \midrule
 ClimaConvo Task B~\cite{shiwakoti2024analyzing}& climate change objectives and related activist movement& $X$ (Twitter)& 15,309 tweets& 2024& English&Support, Neutral, Oppose\\
   % \midrule
 EcoVerse~\cite{grasso-etal-2024-ecoverse-annotated}& environmental topics& $X$ (Twitter)&  3,023 tweets & 2024& English&eco-related / not eco-related; negative, positive, neutral; supportive, neutral, skepti-
cal/opposing\\
   % \midrule
 EZ-STANCE~\cite{anonymous2024ezstance}& Covid Epidemic, World Events, Education and Culture, Entertainment and Consumption, Sports, Rights, Environmental Protection, and Politics& $X$ (Twitter)&  30,606  tweets & 2024& English &Favor, Against, Neutral\\
   % \midrule
GunStance~\cite{gyawali-etal-2024-gunstance}& Regulating guns, Banning guns& $X$ (Twitter)&  5,500  tweets & 2024& English &In-Favor, Against, Neutral\\
   % \midrule
GenderStance~\cite{li-zhang-2024-pro}& Gender Bias & Kialo& 36k sentences(automatic annotation)  & 2024& English &In-Favor, Against, None\\
   \bottomrule
% CHATStance~\cite{zhao-etal-2024-zerostance}& open-domain & ChatGPT & 36k sentences  & 2024& English &In-Favor, Against, None\\
% \hline
  \end{tabular}
}
\end{table*}

\begin{table*}
  \caption{Multi-target stance detection datasets}
  \label{mtsd}
  \resizebox{\linewidth}{!}{
  \begin{tabular}{l>{\raggedright\arraybackslash}p{0.2\linewidth}llll>{\raggedright\arraybackslash}p{0.18\linewidth}}
    \toprule
    \textbf{Dataset-name} & \textbf{Target(s)} & \textbf{Source} & \textbf{Size} & \textbf{Time} & \textbf{Language} &\textbf{Annotation Classes}\\
    % \hline
    \midrule
 Multi-Target Stance~\cite{sobhani-etal-2017-dataset}& \{Clinton-Sanders\}, \{Clinton-Trump\}, \{Cruz-Trump\}& $X$ (Twitter) & 4,455 tweets& 2017& English&Favor, Against, Neither\\
 % \midrule
 Trump-Hillary~\cite{darwish2017trump}& Hillary Clinton,
Donald Trump& $X$ (Twitter) & 3,450 tweets& 2017& English&support Trump, attack Trump, support Clinton, attack Clinton, neutral/irrelevant\\
\bottomrule
  \end{tabular}

}
\end{table*}

\begin{table*}
  \caption{Claim-based stance detection datasets}
  \label{cbdata}
  \resizebox{\linewidth}{!}{
  \begin{tabular}{l>{\raggedright\arraybackslash}p{0.25\linewidth}>{\raggedright\arraybackslash}p{0.1\linewidth}>{\raggedright\arraybackslash}p{0.1\linewidth}ll>{\raggedright\arraybackslash}p{0.18\linewidth}}
\toprule
    \textbf{Dataset-name/Authors}& \textbf{Target(s)} & \textbf{Source} & \textbf{Size} & \textbf{Time} & \textbf{Language} &\textbf{Annotation Classes}\\
\midrule
    Emergent~\cite{ferreira2016emergent}&  Claims and news headlines&snopes.com and $X$ (Twitter)& 300 claims and 2,595 headlines& 2016&   English&For, Against, Observing\\
% \midrule
    % IAC 2.0~\cite{abbott-etal-2016-internet}& Various topics& Online debates& 482 posts& 2016&   English&Pro, Con\\
    PHEME~\cite{zubiaga2016analysing}&Claims& $X$ (Twitter), Reddit& 330 rumour threads (4,842 tweets)& 2017&   English &Rumours, Non-Rumours\\
    RumourEval~\cite{derczynski-etal-2017-semeval}& Rumorous tweets& $X$ (Twitter), Reddit& 5,568 tweets (4,519 + 1,049)& 2017&   English&Support, Deny, Query, Comment\\
% \midrule
    FNC~\cite{pomerleau2017fake}& News headlines& FNC& 49,972 annotated and 25,413 unannotate headline-body pairs& 2017& 
      English&Agrees, Disagrees, Discusses, Unrelated\\
      % \midrule
    
    \citet{rohit2018analysis}& The bill/issue of the speech under
consideration& Indian Parliament& 1,201 speeches& 2018&   English&Favor, Against for stance; Appreciate, Blame, Call for Action, Issue for purpose\\
% \midrule
    \citet{baly-etal-2018-integrating}& Claims extracted from Web sites&  VERIFY \& REUTERS& 402 claims and
3,042 annotated
documents& 2018&  Arabic&Agree, Disagree, Discuss, Unrelated\\
% \midrule

 \citet{lozhnikov2020stance}& Claims extracted from news and
tweets& $X$ (Twitter)  & 700 tweets and
200 news
articles& 2018& Russian&Support, Deny, Query,
Comment\\
% \midrule
    C-STANCE Subtask B~\cite{zhao-etal-2023-c}& Covid Epidemic & Weibo &29,088 tweets & 2023&   Chinese &Favor, Against, Neutral\\
\bottomrule
  \end{tabular}
}
\end{table*}

\begin{table*}
  \caption{Coversational-based stance detection datasets}
  \label{csddata}
  \resizebox{\linewidth}{!}{
  \begin{tabular}{l>{\raggedright\arraybackslash}p{0.2\linewidth}llll>{\raggedright\arraybackslash}p{0.18\linewidth}}
\toprule
    \textbf{Dataset-name/Authors}& \textbf{Target(s)} & \textbf{Source} & \textbf{Size} & \textbf{Time} & \textbf{Language} &\textbf{Annotation Classes}\\
    % \hline 
\midrule
\citet{lai2018stance}& 2016 referendum on reform of the Italian Constitution& $X$ (Twitter)  & 993 triplets (2,889 tweets)& 2018& Italian&Favor, Against, None\\
 % \midrule
 SRQ~\cite{villa2020stance}& Student Marches, Iran Deal, Santa Fe Shooting, General Terms & $X$ (Twitter) & 5,220 tweets& 2020 & English&Comment, Explicit Denial, Implicit Denial, Explicit Support, Implicit Support, Queries\\
 % \midrule
 Cantonese-CSD~\cite{li2022improved} & Covid-19 & Hong Kong Social media& 5,876 items& 2022 & Cantonese&Favor, Against, Neither\\
 % \midrule
 MT-CSD~\cite{niu2024challenge} & Bitcoin, Tesla, SpaceX, Donald Trump, Joe Biden & Reddit & 15,876 items& 2023 & English&Favor, Against, None\\
\bottomrule
  \end{tabular}
}
\end{table*}

\begin{table}
% \captionsetup{font=small}
 % \vspace{3cm}
  % \small
  \resizebox{\linewidth}{!}{
  \begin{tabular}{c|c|c|c|c}
    \hline
    \textbf{Type} & \textbf{Sentence-level} & \multicolumn{3}{c}{\textbf{Conversation-based}}\\
    \hline
    \textbf{Classif. Task} &  Sentence classification & \multicolumn{3}{c}{\makecell[c]{ Conversation history \\classification}} \\
    \hline
    \textbf{Label} & \multicolumn{4}{c}{Favor, Against, None} \\
    \hline
    \textbf{Work} & \makecell[c]{SEM16, P-stance\\ COVID-19-Stance\\ VAST, WT-WT} & SRQ & Cantonese-CSD & \makecell[c]{MT-CSD}\\
    \hline
    \textbf{Target-nums} & $\geq $4  & 4 & 1 & 5\\
    \hline
    % \textbf{Multi-turn} & - & \contour{black}{\times} & \contour{black}{\checkmark}  & \contour{black}{\checkmark}\\
    % \textbf{Multi-turn} & - & $\times$ & $\checkmark$ & $\checkmark$ \\
    \textbf{Multi-turn} & - & $\times$ & $\checkmark$ & $\checkmark$\\
    \hline
    % \textbf{English} & \contour{black}{\checkmark} &\contour{black}{\times} & \contour{black}{\times}  & \contour{black}{\checkmark}\\
   
 \textbf{English}& $\checkmark$& $\checkmark$& $\times$&$\checkmark$\\
  \hline
  \end{tabular}
  }
\caption{ Comparison of different stance detection datasets}
\label{csddatacmp1}
\end{table}

\begin{table*}
  \caption{Multi-modal stance detection datasets}
  \label{mmsdd}
  \resizebox{\linewidth}{!}{
  \begin{tabular}{l>{\raggedright\arraybackslash}p{0.2\linewidth}>{\raggedright\arraybackslash}p{0.2\linewidth}lll>{\raggedright\arraybackslash}p{0.2\linewidth}l}
\toprule
    \textbf{Dataset-name} & \textbf{Target(s)} & \textbf{Source} & \textbf{Size} & \textbf{Time} & \textbf{Language} &\textbf{Annotation Classes} &\textbf{Modal}\\
 \midrule
 Levow et al.~\cite{levow2014recognition}& Decisions on item placement (inventory task) and whether to fund or cut expenses (budget task) in a superstore& Spontaneous speech& $\sim$7.6 hours& 2014& English&No Stance, Weak Stance, Moderate Stance, Strong Stance, Unclear for stance; Positive, Negative, Neutral, Unclear for polarity &aural \\
 % \midrule
    % \hline
    MMVAX-STANCE~\cite{weinzierl-harabagiu-2023-identification}& Covid-19& $X$ (Twitter)  & 4,870 tweets& 2023&  English& Accept, Reject, No Stance, Not Relevant&visual \\
    % \midrule
 MSDD~\cite{DBLP:conf/clsw/HuLWZL22}& Various topics& Friends and The Big Bang Theory& 1,296 video & 2023& English& Postitive, Negative, Neutral&visual, aural\\
    % \midrule
    MMSD~\cite{liang2024multi}& Twitter Stance Election, COVID-CQ, MWTWT, Russo-Ukrainian Conflict, Taiwan Question,  & $X$ (Twitter) & 17,544 tweets& 2024 & English& Favor, Against, Neutral&visual \\
    % \midrule
    MmMtCSD~\cite{niu2024multimodal}& Bitcoin, Tesla, Post-T & Reddit & 21,340 items & 2024 &  English& Favor, Against, None& visual \\
\bottomrule
  \end{tabular}
}
\end{table*}

\begin{table*}
  \caption{Multilingual stance detection datasets}
  \label{Multilingual}
  \resizebox{\linewidth}{!}{
  \begin{tabular}{l>{\raggedright\arraybackslash}p{0.2\linewidth}l>{\raggedright\arraybackslash}p{0.1\linewidth}lll}
\toprule
    \textbf{Dataset-name/Authors}& \textbf{Target(s)} & \textbf{Source} & \textbf{Size} & \textbf{Time} & \textbf{Language} &\textbf{Annotation Classes}\\
\midrule
    IberEval-2017~\cite{taule2017overview}& Independence of Catalonia& $X$ (Twitter) & 5,400 tweets in Spanish and 5,400 tweets in Catalan& 2017&   Catalan and Spanish&Favor, Against, None\\
    % \midrule
    \citet{swami2018englishhindi}& Demonetisation in India in 2016& $X$ (Twitter) & 3,545 tweets& 2018&   English-Hindi&Favor, Against, None\\
    % \midrule
    MultiStanceCat~\cite{taule2018overview}& Catalan Referendum& $X$ (Twitter)& 11,398 tweets& 2018&   Catalan
and Spanish&Favor, Against, Neither\\
    % \midrule
X-Stance~\cite{vamvas2020xstance}& 175 communal, cantonal and national elections between 2011 and 2020& Smartvote& 67,000 comments& 2020& German and French&Favor, Against\\
\bottomrule
  \end{tabular}
}
\end{table*}

\textbf{Multi-target stance detection.}
Building on the foundation of target-specific stance detection, the multi-target stance detection task addresses the scenario where a single tweet may contain multiple related targets. This task involves predicting the stance polarity of multiple targets within the same tweet (or statement). The concept of multi-target stance detection is commonly used to analyze the relationships between multiple targets, which often exhibit strong correlations, such as multiple candidates in a political election \cite{lai2018stance, darwish2017trump, wei2018multi}.

\textbf{Claim-based stance detection.} 
In contrast to target-specific stance detection, which focuses on explicit entities or topics, claim-based stance detection involves analyzing the stance towards a specific claim or news article. The primary objective is to determine the stance of comments responding to the news, whether they confirm or challenge the claim's validity. The prediction labels typically take the form of confirming or denying the claim. Claim-based stance detection is a crucial approach for evaluating the veracity of misinformation. By leveraging the stance of replies to predict the claim's veracity ~\cite{ferreira2016emergent}, this method has been recognized as a key technique for rumor and fake news detection ~\cite{ZubiagaABLP18}, highlighting its importance in assessing the credibility of online information.

\textbf{Conversational stance detection.}
In the domain of social media analysis, users frequently express their opinions within conversational threads. However, traditional stance detection techniques often struggle to accurately identify stances in such dialogic environments due to their limited ability to understand contextual nuances. To address this limitation, coversational-based stance detection tasks have been proposed \cite{li2022improved}. Typically, these tasks involve predicting the stance polarity of a given tweet (or statement) based on the conversation history, the current tweet (or statement), and the stance target.

\textbf{Multi-modal stance detection.}
With the rapid evolution of social media, users increasingly express their opinions not only through text, but also through diverse multimedia content, such as audio, images, videos, and emojis. To address this trend, multi-modal stance detection has emerged as a research area, aiming to predict stance polarity by integrating various data forms, including text, images, videos, and audio.

\textbf{User-level stance detection.}
The user-level stance detection task involves predicting a user's stance towards a given topic. In this context, various user attributes can be integrated with the text of their posts to enhance the predictive model~\cite{aldayel2021stance, lynn-etal-2019-tweet}. Notably, most studies in this area have adopted a binary classification approach, categorizing users' stances into two polarized classes: support and against~\cite{darwish2017improved}. This simplification is often justified by the assumption that users will exhibit a clear stance, even if their posts appear neutral.

\subsection{Method Development Perspective}
From a method development perspective, this section provides a detailed introduction to several task categorization approaches.

\textbf{In-target stance detection.}
In the early stages of stance detection research, the in-target stance detection task was the most extensively studied task. This task is closely related to the development of text classification techniques. The in-target stance detection task involves training and inferring stance detection models on the same target. Over the years, a large body of work has focused on this task, exploring various approaches and methodologies.
The task is formally defined as:

\begin{center}
    \fcolorbox{black}{blue!10}{\parbox{0.92\linewidth}{ Let $X={\{x_i,q_i\}}_{i=1}^N$ represent the labeled data collection, where $x$ refers to the input text and $q$ corresponds to the target. $N$ denotes the total number of instances in $X$.  Each sentence-target pair $(x,q) \in X$ is assigned a stance label $y$. In-target stance detection aims to predict the stance label for $x$ towards target $q$. 
    }}
\end{center}

\textbf{Cross-target stance detection.}
In real-world applications, obtaining rich annotated data for each target is a time-consuming and labor-intensive process. Conventional stance detection methods struggle to generalize well across targets. Cross-target stance detection \citep{wei2019modeling} aims to infer the attitude towards a target by leveraging a large amount of annotated data from another target. Typically, most works in this area focus on targets within the same domain, such as Trump and Hillary in the politics domain. However, some studies also explore cross-domain stance detection, where the source and target domains differ, such as training on Trump data and inferring stance on feminist movement data. This approach enables the model to generalize stance detection across different domains and contexts.
The task is formally defined as:

\begin{center}
    \fcolorbox{black}{blue!10}{\parbox{0.92\linewidth}{ Let $X^{s} = {(x^{s}_{i}, q^{s}_{i})}_{i=1}^{N^{s}}$ denote the labeled dataset in the source domain, where each tuple $(x^{s}_{i}, q^{s}_{i})$ consists of an input text $x^{s}_{i}$ and its corresponding target $q^{s}_{i}$. The dataset contains $N^{s}$ instances, each associated with a stance label $y^{s}$. Given an input sentence $x^{d}$ and a target $q^{d}$ in the target domain, the goal of cross-target stance detection is to predict the stance label for $x^{d}$ towards $q^{d}$ by leveraging the model learned from the labeled data $X^{s}$ in the source domain. 
    }}
\end{center}

\textbf{Zero-shot stance detection.}
In real-world applications, test data often involves unseen targets during training. As a special case of cross-target stance detection, the zero-shot stance detection Task involves training a model on annotated data from a specific target and directly predicting the stance on unseen targets. Pioneering work by \citet{allaway2020zero} first introduced the zero-shot stance detection task and proposed the VAST dataset, which contains a large number of targets with no overlap between training and testing targets. Furthermore, some studies have begun to explore open-domain zero-shot stance detection, aiming to simulate real-world scenarios and reduce the need for training samples.
The task is formally defined as:

\begin{center}
    \fcolorbox{black}{blue!10}{\parbox{0.92\linewidth}{  
Let $X^{s} = {(x^{s}_{i}, q^{s}_{i})}_{i=1}^{N^{s}}$ denote the labeled dataset in the source domain, where each tuple $(x^{s}_{i}, q^{s}_{i})$ consists of an input text $x^{s}_{i}$ and its corresponding target $q^{s}_{i}$.
$X^{d} = {(x^{d}_{i}, q^{d}_{i})}_{i=1}^{N^{d}}$ denote a set of unlabeled instances towards unseen or testing targets.  Note that there is no overlap between $X^{s}$ and $X^{d}$.
The aim of zero-shot stance detection is to train a model from each sentence $x^{s}_{i}$ towards the source known target $q^{s}_{i}$ from $X^{s}$, and infer the stance polarity for each
sentence $x^{d}_{i}$ towards an unseen target $q^{d}_{i}$ in $X^{d}$.
}
    }
\end{center}

\textbf{Cross-lingual stance detection.}
In today's globalized world, understanding and analyzing stances in multiple languages is crucial, particularly given the linguistic diversity of online platforms such as social media, forums, and news websites. However, traditional monolingual stance detection methods often fall short in multilingual applications due to language and cultural variations. This has led to the development of multilingual stance detection tasks, which aim to expand the scope and applicability of stance detection research ~\cite{swami2018englishhindi}.
The task is formally defined as:

\begin{center}
\fcolorbox{black}{blue!10}{\parbox{0.92\linewidth}{
Let $X^{s} = {(x^{s}_{i}, q^{s}_{i}, l^{s}_{i})}_{i=1}^{N^{s}}$ denote the labeled dataset in the source language, where each tuple $(x^{s}_{i}, q^{s}_{i}, l^{s}_{i})$ consists of an input text $x^{s}_{i}$, its corresponding target $q^{s}_{i}$, and the language label $l^{s}_{i}$.
$X^{t} = {(x^{t}_{i}, q^{t}_{i}, l^{t}_{i})}_{i=1}^{N^{t}}$ denote a set of unlabeled instances in the target language. Note that there is no overlap between $X^{s}$ and $X^{t}$, and $l^{s}_{i} \neq l^{t}_{i}$.
The aim of cross-lingual stance detection is to train a model from each sentence $(x^{s}_{i}$ towards the source known target $q^{s}_{i}$ from $X^{s}$, and infer the stance polarity for each sentence $x^{t}_{i}$ towards an unseen target $q^{t}_{i}$ in $X^{t}$, despite the language difference between $l^{s}_{i}$ and $l^{t}_{i}$.
}
}
\end{center}

\section{Stance detection Datasets}
This section provides a comprehensive overview of existing stance detection task datasets, presented in the order of their release.

\subsection{Target-Specific Datasets}
As discussed earlier, target-specific stance detection is the most widely studied task in this field. Therefore, a large number of datasets have been made available, including multi-modal and multilingual resources.

Among these, the SemEval stance detection dataset SemEval-2016 task 6~\cite{MohammadKSZC16} is the most widely used, providing two sub-datasets to support both supervised (Task A) and weakly supervised (Task B) frameworks. It contains 4,870 labeled tweets.
NLPCC-2016 is the first widely used Chinese benchmark dataset, collected from Weibo. It contains 4,000 labeled sentences \cite{xu2016overview}.
P-stance dataset~\cite{li2021p} was introduced for the political domain, featuring three targets (Donald Trump, Joe Biden, and Bernie Sanders) and collected during the 2020 US presidential election. 
The WT-WT dataset~\cite{conforti2020will} is one of the largest manually annotated datasets, covering multiple domains and containing 51,284 annotated samples. The COVID-19-Stance dataset~\cite{glandt2021stance} focuses on the public health domain, with targets such as Anthony S. Fauci, M.D., and policies like Keeping Schools Closed and Wearing a Face Mask, comprising 7,122 samples. 
Furthermore, the EZ-STANCE dataset~\cite{anonymous2024ezstance} expands the scope to various domains, including Covid Epidemic, World Events, Education and Culture, Entertainment and Consumption, Sports, Rights, Environmental Protection, and Politics, with 30,606 annotated samples. A more detailed description of these datasets can be found in Table~\ref{intar-dataset}.

\subsection{Multi-Targets Datasets}
Multi-target stance detection datasets primarily focus on the political domain, such as the US presidential election. Two notable datasets are the Multi-targets dataset~\cite{sobhani-etal-2017-dataset} and the Trump vs. Hillary dataset~\cite{darwish2017trump}. The Multi-targets dataset involves annotating each tweet with stance labels for both candidates simultaneously, with target pairs including Clinton-Sanders, Clinton-Trump, and Cruz-Trump. The labels used are Favor, Against, and Neither. Similarly, the Trump vs. Hillary dataset annotates each tweet with stance labels for both candidates, such as supporting Hillary and opposing Trump.
A more detailed description of these datasets can be found in Table~\ref{mtsd}.

\subsection{Claim-based Datasets}
In stance detection datasets, the target is typically a single statement, such as a tweet post. 
The Emergent dataset~\cite{ferreira2016emergent} consists of claims extracted from rumor sites and $X$ (Twitter), comprising 300 claims and 2,595 corresponding headlines.
The RumourEval dataset~\cite{derczynski-etal-2017-semeval} is a claim-based stance detection dataset designed for rumor resolution, comprising 5,568 tweets from $X$ platform and Reddit. Each rumor tweet is evaluated against a set of other tweets to determine the stance of these tweets in supporting or denying the source of the rumor.
Additionally, the FNC dataset~\cite{pomerleau2017fake} targets news headlines, with 49,972 annotated labels and 25,413 unannotated headline-body pairs. 
\citet{rohit2018analysis} extracted 1,200 samples from the Indian Parliament dataset, focusing on the target of ``The bill/issue of the speech under consideration''.
Multilingual datasets have also been developed, such as the SemEval 2019 rumors detection dataset~\cite{li2019eventai}, which extends the SemEval 2017 dataset by adding new data from Reddit and incorporating Russian as a new topic.
\citet{baly-etal-2018-integrating} developed the Russian dataset, and C-STANCE Subtask B is the Chinese dataset collected from Weibo.
A more detailed description of these datasets can be found in Table~\ref{cbdata}.

\subsection{Coversational-based Datasets}
Recent years have seen a growing interest in conversational stance detection datasets focused on social media comment data. One of the first datasets is proposed by \citet{lai2018stance}, which explores the 2016 referendum on the reform of the Italian Constitution. 
The SRQ dataset~\cite{villa2020stance}, collected from the $X$ platform, contains 5,220 tweets with a primary focus on single-round reply conversations. The Cantonese-CSD dataset~\cite{li2022improved}, sourced from a Hong Kong social media platform, targets Covid-related discussions in Cantonese. The MT-CSD dataset~\cite{niu2024challenge} is currently the largest and most in-depth conversational stance detection dataset, featuring multiple targets such as ``Bitcoin, Tesla, SpaceX, Donald Trump, and Joe Biden'', with a total of 15,876 annotated samples.
A detailed description of the dataset is provided in Table~\ref{csddata}. 
Additionally, we present a comparison between conversation-level and sentence-level stance detection in Table~\ref{csddatacmp1}.

\begin{table*}
  \caption{User-level stance detection datasets}
  \label{Userlevel}
  \resizebox{\linewidth}{!}{
  \begin{tabular}{l>{\raggedright\arraybackslash}p{0.2\linewidth}l>{\raggedright\arraybackslash}p{0.1\linewidth}l>{\raggedright\arraybackslash}p{0.1\linewidth}l}
\toprule
    \textbf{Dataset-name/Authors} & \textbf{Target(s)} & \textbf{Source} & \textbf{Size} & \textbf{Time} & \textbf{Language} &\textbf{Annotation Classes}\\
\midrule
   \citet{DBLP:conf/icwsm/DarwishSAN20} &Kavanaugh, Trump, Erdogan &$X$ (Twitter) &72690 user (automatic annotation) &2020& Kavanaugh and Trump dataset (English), Erdogan  dataset (Turkish) &pro, anti\\
    % \midrule
\citet{DBLP:conf/rev/ElzanfalyRO22} & tweet and article &$X$ (Twitter) &32,905 user&2022& English, &Unsupervised task \\
    % \midrule
    % DoubleH
    \citet{DBLP:conf/icwsm/ZhangZP024} &Joe Biden, Donald Trump &$X$ (Twitter) & 1,123,749 tweets (automatic annotation) &2024 &English&pro-Trump, pro-Biden\\
\bottomrule
  \end{tabular}
}
\end{table*}

\subsection{Multi-Modal Datasets}
The earliest multi-modal stance detection dataset, proposed by Levow et al.~\cite{levow2014recognition}, focused on spontaneous speech audio data, with targets including decisions on item placement and budget allocation in a superstore. 
The MMVAX-STANCE dataset~\cite{weinzierl-harabagiu-2023-identification}, comprising 4,870 tweets, targets Covid-19 and incorporates visual information. The MSDD dataset~\cite{DBLP:conf/clsw/HuLWZL22}, sourced from TV shows Friends and The Big Bang Theory, contains 1,296 video clips with multiple targets. The MMSD dataset~\cite{liang2024multi}, collected from $X$ platform, includes 17,544 tweets with targets such as $X$ (Twitter) Stance Election, COVID-CQ, MWTWT, Russo-Ukrainian Conflict, and Taiwan Question. The MmMtCSD dataset~\cite{niu2024multimodal} is the first multi-modal, conversational stance detection dataset, featuring text, image, and video modalities, with 21,340 items. Detailed information is presented in Table~\ref{mmsdd}.
% $\mathcal{X}$

\subsection{Multilingual Datasets}
In the previous section, we introduced several datasets from different languages. This section focuses on datasets that encompass multiple languages for stance detection tasks.
For instance, the IberEval-2017 dataset~\cite{taule2017overview} contains 5,400 tweets in Spanish and 5,400 tweets in Catalan, with the goal of determining the independence of Catalonia.
\citet{swami2018englishhindi} introduced an English-Hindi language dataset consisting of 3,545 tweets. 
The MultiStanceCat dataset~\cite{taule2018overview} is annotated with the Catalan Referendum as the target, comprising 11,398 tweets from 2018 in both Catalan and Spanish languages. The X-Stance dataset~\cite{vamvas2020xstance} is a German and French dataset, consisting of 67,000 comments. Detailed information is provided in Table~\ref{Multilingual}.

\subsection{User-level Datasets}
Due to the challenges of manual annotation, user stance detection datasets often employ automatic annotation methods, such as leveraging hashtags from social media platforms. For instance, 
\citet{DBLP:conf/icwsm/DarwishSAN20} utilized automatic annotation to label 72,690 users with respect to their stances on Kavanaugh, Trump, and Erdogan. 
\citet{DBLP:conf/rev/ElzanfalyRO22} proposed an unsupervised task involving 32,905 users. 
\citet{DBLP:conf/icwsm/ZhangZP024} constructed a dataset in the political domain, focusing on Joe Biden and Donald Trump as targets, and employed automatic annotation to label 1,123,749 tweets. The details of these datasets are presented in Table~\ref{Userlevel}.

\begin{figure*}[h]
  \centering 
  \includegraphics[width=\linewidth]{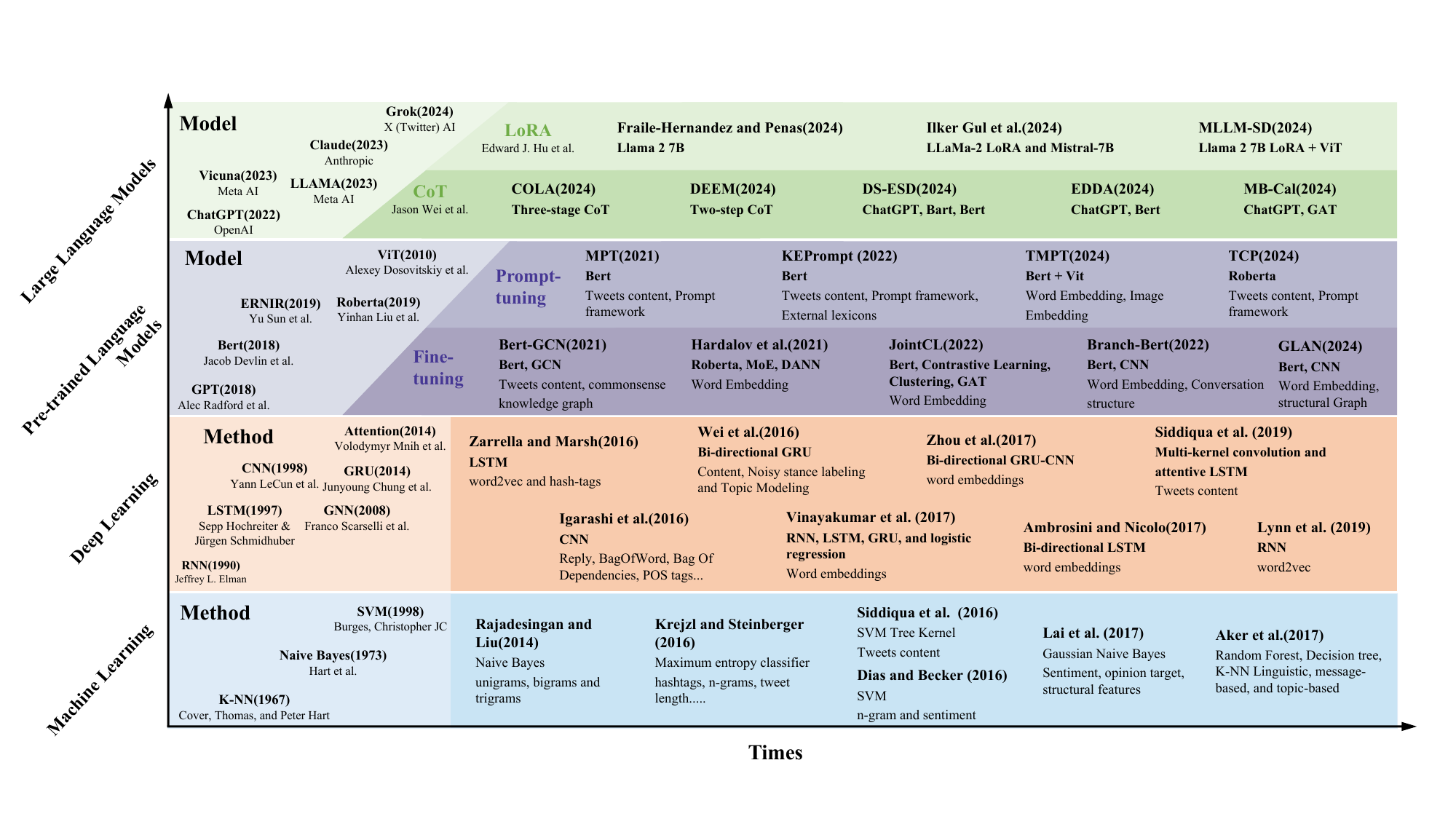}
\caption{The taxonomy of stance detection methods.} 
\label{methods}
\end{figure*}

\section{Traditional Stance Detection Algorithms}

As mentioned earlier, the development of stance detection models is closely related to the advancement of text classification models. Various approaches have been employed to model and train stance classifiers. Fig. \ref{methods} presents the evolution of methods. In this section, we review traditional stance detection methods, including early machine learning-based approaches, conventional deep learning methods, fine-tuning methods based on pre-trained models, and prompt-based fine-tuning methods for stance detection.

\subsection{Feature-Based Machine Learning Approaches}
\begin{figure}[]
\centering
\subfigure[Feature-Based Machine Learning Approaches]{
\includegraphics[width=3.2cm]{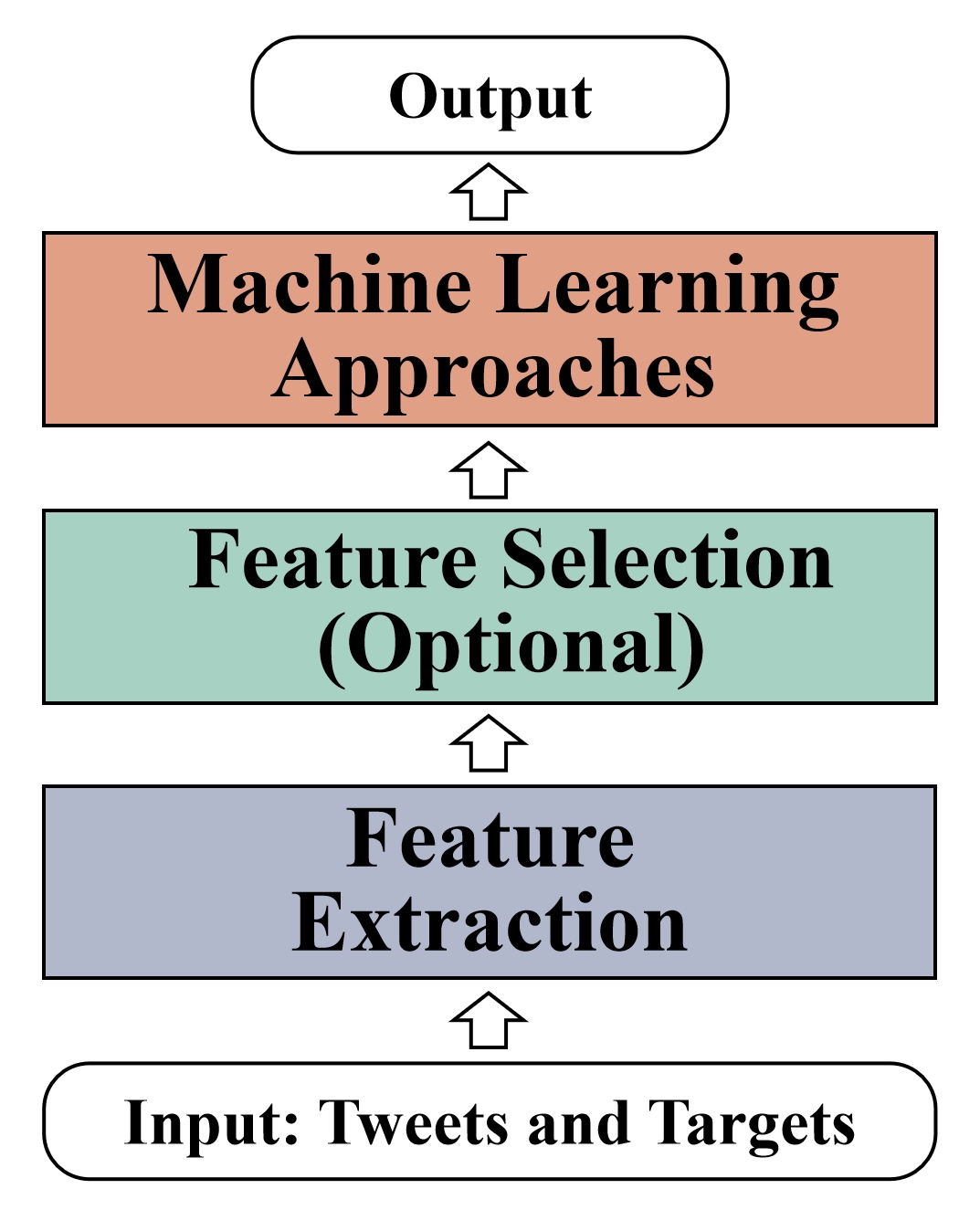}
%\caption{fig1}
}
\quad
\subfigure[Traditional Deep Learning Approaches.]{
\includegraphics[width=3.9cm]{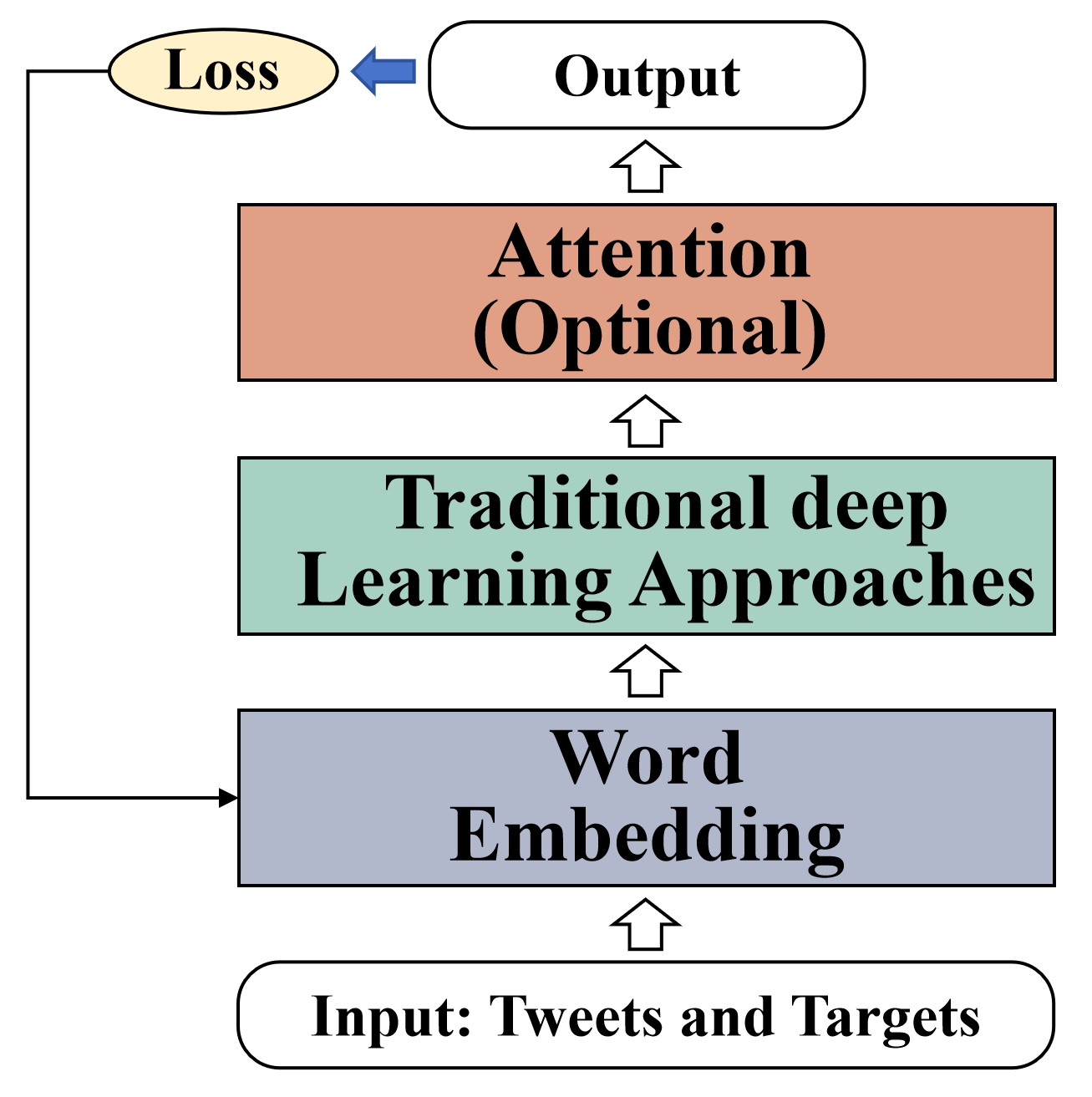}
}
\caption{Framework of traditional methods.}
\label{MLmethods}
\end{figure}

Early stance detection methods primarily relied on traditional machine learning approaches. Their basic framework is illustrated in Fig.~\ref{MLmethods} (a). 
First, a feature identifier is constructed to recognize important features, followed by the development of a traditional machine learning model to achieve classification and prediction. The mainstream methods can be broadly categorized into several types, including Support Vector Machines (SVMs), logistic regression, decision trees, and k-Nearest Neighbors (KNN).

Among these methods, SVMs are the most widely employed approach for stance detection. They have been used in over 40 studies, either as the primary approach or as a baseline for comparison with other methods. Notable examples of studies that have utilized SVMs for stance detection include \cite{addawood2017stance}, \cite{DBLP:conf/eacl/SlonimBSBD17}, and \cite{DBLP:journals/corr/Kucuk17aa}.

Logistic regression is another popular approach for stance detection, appearing in over 15 studies. This method has been used in various forms, including as a standalone classifier and as part of an ensemble classifier. Examples of studies that have employed logistic regression for stance detection include \cite{ferreira2016emergent}, \cite{hacohen2017stance}, and \cite{tsakalidis2018nowcasting}.

Decision trees and random forests are also commonly used approaches for stance detection. Decision trees have been employed in several studies, including \cite{addawood2017stance} and \cite{simaki2017stance}. Random forests, which are ensemble classifiers based on decision trees, have been used in studies such as \cite{küçük2017stance} and \cite{wojatzki2016stance}.

Finally, artificial neural networks (ANNs) have also been applied to stance detection tasks. Examples of studies that have used ANNs for stance detection include \cite{DBLP:conf/comad/SenSMR18} and \cite{tsakalidis2018nowcasting}. Specifically, classifiers based on Multilayer Perceptron (MLP) have been successfully applied to stance detection tasks in studies such as \cite{rajendran2018stance} and \cite{DBLP:conf/www/ZhangYL18}.
Detailed information is provided in Table~\ref{MLms}.

\begin{table*}
  \caption{Feature-based machine learning approaches}
  \label{MLms}
  \resizebox{\linewidth}{!}{
  \begin{tabular}{l>{\raggedright\arraybackslash}p{0.1\linewidth}>{\raggedright\arraybackslash}p{0.3\linewidth}>{\raggedright\arraybackslash}p{0.25\linewidth}l}
\toprule
    \textbf{Study}& \textbf{Task}& \textbf{Features}& \textbf{Approach}& \textbf{Dataset}\\
\midrule
    \citet{rajadesingan2014identifying}& target-specific&unigrams, bigrams and trigrams & Naive Bayes &Debate\\
    \citet{dias2016inf}& target-specific& n-gram and sentiment & SVM & SemEval-2016\\
    \citet{krejzl2016uwb}& target-specific& hashtags, n-grams, tweet length, Part-of-speech, General Inquirer, entity-centered sentiment dictionaries, Domain Stance Dictionary & Maximum entropy classifier & SemEval-2016\\
    \citet{ebrahimi2016joint}& target-specific& n-gram and sentiments & Discriminative and generative models & SemEval-2016\\
    \citet{elfardy2016cu}& target-specific & Lexical Features, Latent Semantics, Sentiment, Linguistic Inquiry, Word Count and Frame Semantics features & SVM & SemEval-2016\\
    \citet{siddiqua2018stance}& target-specific& Tweets content & SVM Tree Kernel & SemEval-2016\\
    \citet{lai2017friends}& target-specific & Sentiment, opinion target, structural features (hashtags, mentions, punctuation marks), text-Based features & Gaussian Naive Bayes classifier & SemEval-2016\\
    \citet{10.1145/3359307}& target-specific& Networks features& SVM & SemEval-2016\\
    \citet{aker2017simple}& claim-based& Linguistic, message-based, and topic-based such as (Bag of words, POS tag, Sentiment, Named entity and others& Random Forest, Decision tree and Instance Based classifier (K-NN) & RumourEval, PHEME\\
\bottomrule
  \end{tabular}
}
\end{table*}

\begin{table*}
  \caption{Traditional deep learning approaches}
  \label{DLm}
  \resizebox{\linewidth}{!}{
  \begin{tabular}{l>{\raggedright\arraybackslash}p{0.1\linewidth}>{\raggedright\arraybackslash}p{0.3\linewidth}>{\raggedright\arraybackslash}p{0.25\linewidth}l}
\toprule
    \textbf{Study}& \textbf{Task}& \textbf{Features}& \textbf{Approach}& \textbf{Dataset}\\
    
\midrule
    \citet{zarrella-marsh-2016-mitre}& target-specific& word2vec and hash-tags& LSTM& SemEval-2016\\
% \midrule
 \citet{wei2019topic}& target-specific& Content and Sentiment lexicon& BiLSTM&SemEval-2016\\
 
    \citet{igarashi-etal-2016-tohoku}& target-specific& Reply, BagOfWord, BagOfDependencies, POS tags Sentiment WordNet, Sentiment Word Subject, Target Sentiment and Point-wise Mutual Information& CNN& SemEval-2016\\
% \midrule
 \citet{kochkina2017turing}& claim-based& word2vec, Tweet lexicon, Punctuation, Attachments,Relation to other tweets, Content length and Tweet role.& Branch-LSTM&RumourEval \\
 % \midrule
 % \midrule
 \citet{sobhani-etal-2017-dataset}& multi-target& word vectors& Bidirectional RNN&Multi-Target Stance\\
 % \midrule
 % \midrule
 \citet{wei-etal-2016-pkudblab}& target-specific& Content, Noisy stance labeling and Topic Modeling& BiGRU&SemEval-2016\\
 % \midrule
 \citet{zhou2017connecting}& target-specific& word embeddings& Bi-directional GRU-CNN&SemEval-2016\\
    % \midrule
    % \midrul
    \citet{vinayakumar2017deep}& multilingual& Word embeddings& RNN, LSTM, GRU, and logistic regression& MultiStanceCat\\
% \midrule
    \citet{ambrosini2017neural}& multilingual& Word embeddings& LSTM, bidirectional LSTM, CNN& MultiStanceCat\\
    
 \citet{siddiqua2019tweet}& multi-target& Tweets content& Multi-kernel convolution and
attentive LSTM&Multi-Target Stance\\
 \citet{lynn-etal-2019-tweet}& target-specific& word2vec& RNN&SemEval-2016\\
 SEKT~\cite{bowenacl}& target-specific& Tweets content, ,semantic and emotion lexicons& LSTM + GCN&SemEval-2016\\
\bottomrule
  \end{tabular}
}
\end{table*}

\begin{table*}
  \caption{Pre-trained language models (PLMs)}
  \label{PLMM}
  \resizebox{\linewidth}{!}{
  \begin{tabular}{l>{\raggedright\arraybackslash}p{0.15\linewidth}>{\raggedright\arraybackslash}p{0.25\linewidth}>{\raggedright\arraybackslash}p{0.28\linewidth}>{\raggedright\arraybackslash}p{0.18\linewidth}}
\toprule
    \textbf{Study}& \textbf{Task}& \textbf{Features}& \textbf{Approach}& \textbf{Dataset}\\
\midrule
 % Bert-Joint~\cite{bert}& target-specific& Tweets content, topics information& Bert&SemEval-2016\\
% \midrule
 Bert-GCN~\cite{liu2021enhancing}& target-specific& Tweets content, commonsense knowledge graph& Bert + GCN&VAST\\
 
    \citet{li2021p}& target-specific& Word Embedding& BERTweet(Pre-train Bert)& P-stance\\

\citet{hardalov2021cross}& target-specific& Word Embedding
& Roberta + MoE + DANN
& SemEval-2016, VAST, WT-WT\\

 TarBK~\cite{zhu2022enhancing}& target-specific& Tweets content, targeted background knowledge from Wikipedia& Bert&SemEval-2016, VAST, WT-WT\\
% \midrule
 % SEKT~\cite{zhu2022enhancing}& target-specific& Tweets content, ,semantic and emotion lexicons& Bert + GCN&SemEval-2016\\
 
WS-BERT~\cite{he2022infusing}& target-specific& Word Embedding, Encoding Wikipedia Knowledge& Bert + BertTweet
& COVID-19-Stance, P-stance\\

 JointCL~\cite{liang2022jointcl}& target-specific& Word Embedding
& Bert + Contrastive Learning + Clustering + GAT&SemEval-2016, VAST, WT-WT\\

Branch-Bert~\cite{li2022improved}&Coversational-based & Word Embedding, Conversation structure&   Bert + CNN&Cantonese-CSD\\
 
 \citet{zhang2024commonsense}& target-specific& Word Embedding, Knowledge Graph& RoBERTa + bidirectional conditional encoding layer + Node2Vec&SemEval-2016, VAST\\

  GLAN~\cite{niu2024challenge}& Coversational-based& Word Embedding,  structural Graph& Bert + CNN + GCN&MT-CSD\\

MSFR~\cite{10446704} & target-specific &Tweets content& Bert & WT-WT,VAST\\

 MPT~\cite{hu2021knowledgeable}& target-specific& Tweets content, prompt framework&   Bert+prompt learning&SemEval-2016, VAST\\

KEPrompt~\cite{huang2023knowledge}& target-specific& Tweets content, prompt framework, external lexicons& Bert+prompt  learning&SemEval-2016, VAST\\

 TMPT~\cite{liang2024multi}& Multi-modal& Word Embedding, Image Embedding& Bert + Vit + prompt learning&MMSD\\

 TCP~\cite{wang-pan-2024-target-adaptive}& target-specific& Tweets content, prompt framework& RoBERTa + prompt learning&SemEval-2016, P-stance\\

\bottomrule
  \end{tabular}
}
\end{table*}

\begin{table*}
  \caption{Approaches based on large language models}
  \label{LLMmethods}
  \resizebox{\linewidth}{!}{
  \begin{tabular}{>{\raggedright\arraybackslash}p{0.18\linewidth}>{\raggedright\arraybackslash}p{0.1\linewidth}>{\raggedright\arraybackslash}p{0.25\linewidth}l}
\toprule
    \textbf{Study}& \textbf{Task}& \textbf{Approach}& \textbf{Dataset}\\
    % \hline
     \midrule
    KASD~\cite{li2023stance}& target-specific& LLM(CoT) + Bert& SemEval-2016, P-Stance, COVID-19-Stance, VAST\\
    EDDA~\cite{ding2024edda}& target-specific& LLM(CoT) + Bert& SemEval-2016, VAST\\
     % \midrule
 COLA~\cite{lan2023stance}& target-specific& three-stage LLM (CoT)&SemEval-2016, P-Stance, VAST\\
     % \midrule
    MB-Cal~\cite{li2024mitigating}& target-specific& LLM(CoT) + GAT& SemEval-2016, VAST\\
     % \midrule
 DS-ESD~\cite{ding2024distantly}& target-specific& LLM(CoT) + Bart +Bert&SemEval-2016, COVID-19-Stance, VAST\\
     % \midrule
 SQBC~\cite{wagner2024sqbc}& multilingual& LLM (CoT) + Bert&X-Stance\\
     % \midrule
    \citet{fraile2024hamison}& target-specific& Llama 2 7B LoRA& ClimaConvo Task B \\
    % \midrule
 \citet{gül2024stance}& target-specific& LLaMa-2 LoRA and Mistral-7B LoRA&SemEval-2016\\
    % \midrule
 DEEM~\cite{wang2024deem}& multi-target& two-step LLM&P-stance, SemEval-2016,  Multi-Target Stance\\
% \midrule
 \citet{taranukhin-etal-2024-stance-reasoner}& target-specific& LLM(CoT)&SemEval-2016, WT-WT, COVID-19-Stance\\
  KAI~\cite{zhang2024knowledge}& target-specific & LLM(CoT) + Bert&SemEval-2016, P-Stance, VAST\\
  MPPT~\cite{ding2024cross1}& target-specific & LLM(CoT) + Bert + prompt learning&SemEval-2016, VAST\\
  
  LKI-BART~\cite{zhang2024llm}& target-specific & LLM(CoT) + Bart& P-Stance, VAST\\
% \midrule
 MLLM-SD~\cite{niu2024multimodal}& Multi-modal  &Llama 2 7B LoRA + ViT & MmMtCSD\\
% \midrule
TR-ZSSD~\cite{weinzierl-harabagiu-2024-tree}& target-specific &LLM(CoT) &SemEval-2016,  CoVaxFrames, MMVax-Stance\\
\bottomrule
  \end{tabular}
}                                                                
\end{table*}

% \begin{figure}
% \centering
% \includegraphics[width=0.3\linewidth]{FIG/ML.png}
% \caption{Framework of feature-based machine learning approaches.} 
% \label{MLmethods}
% \end{figure}

% \begin{figure}
% \centering
% \includegraphics[width=0.4\linewidth]{FIG/DL.png}
% \caption{Framework of traditional deep neural network approaches.} 
% \label{DLmethods}
% \end{figure}

% \begin{figure}
% \centering
% \includegraphics[width=1\linewidth]{FIG/CNNsample.png}
% \caption{CNNsample.} 
% \label{DLmethods}
% \end{figure}

\subsection{Traditional Deep Learning Approaches}
The development of deep learning techniques in stance detection has been marked by several significant advancements, each contributing uniquely to the field. 

One notable example is the Long Short-Term Memory (LSTM) network, introduced by Hochreiter and Schmidhuber in 1997, which has been widely adopted in over ten studies. The LSTM's ability to capture long-term dependencies in text data has consistently yielded superior results, making it a preferred choice for stance detection tasks.
Following the LSTM, Convolutional Neural Networks (CNNs) have also been extensively employed, with notable contributions from studies including \citet{zarrella-marsh-2016-mitre,sobhani-etal-2017-dataset,kochkina2017turing,lynn-etal-2019-tweet}. 
CNNs have proven effective in extracting spatial hierarchies of features, making them well-suited for processing structured text data~\cite{igarashi-etal-2016-tohoku}.
The advent of attention mechanisms has further enhanced the performance of stance detection models, mitigating issues such as the LSTM's limited long-term memory capacity. Attention-based approaches have garnered significant attention, with notable examples including the combination of LSTM and attention mechanisms, as well as CNNs with attention mechanisms~\cite{siddiqua2019tweet,wei2019topic}. The attention mechanism has increased the association between stance targets and informative words or phrases, leading to improved accuracy in stance detection.

Furthermore, text classification methods based on Graph Convolutional Networks (GCNs) have also contributed to the advancement of stance detection~\cite{bowenacl,LiangF00DHX21,Zhiwei2023}. The graph-based approach can effectively integrate syntactic tree structures and contextual knowledge of text and stance targets, thereby enhancing the performance of stance detection models.
% The basic structure is shown in Fig.~\ref{MLmethods} (b).

As illustrated in Fig.~\ref{MLmethods} (b), the overall framework of these works can be divided into three primary steps:
(1) Vectorization of text and target features: first, the input text and target features are converted into vectors using word embedding techniques, such as GloVe~\cite{pennington2014glove} or Word2Vec~\cite{DBLP:journals/corr/abs-1301-3781}.
(2) Construction of deep learning stance models: next, a deep learning model is built to analyze the stance of the text.
(3) Design of loss function and model training: Finally, an appropriate loss function is designed, and the model is trained and optimized through backpropagation using a large number of annotated samples.
A list of specific methods is provided in Table~\ref{DLm}.

\begin{figure}
\centering
\includegraphics[width=0.7\linewidth]{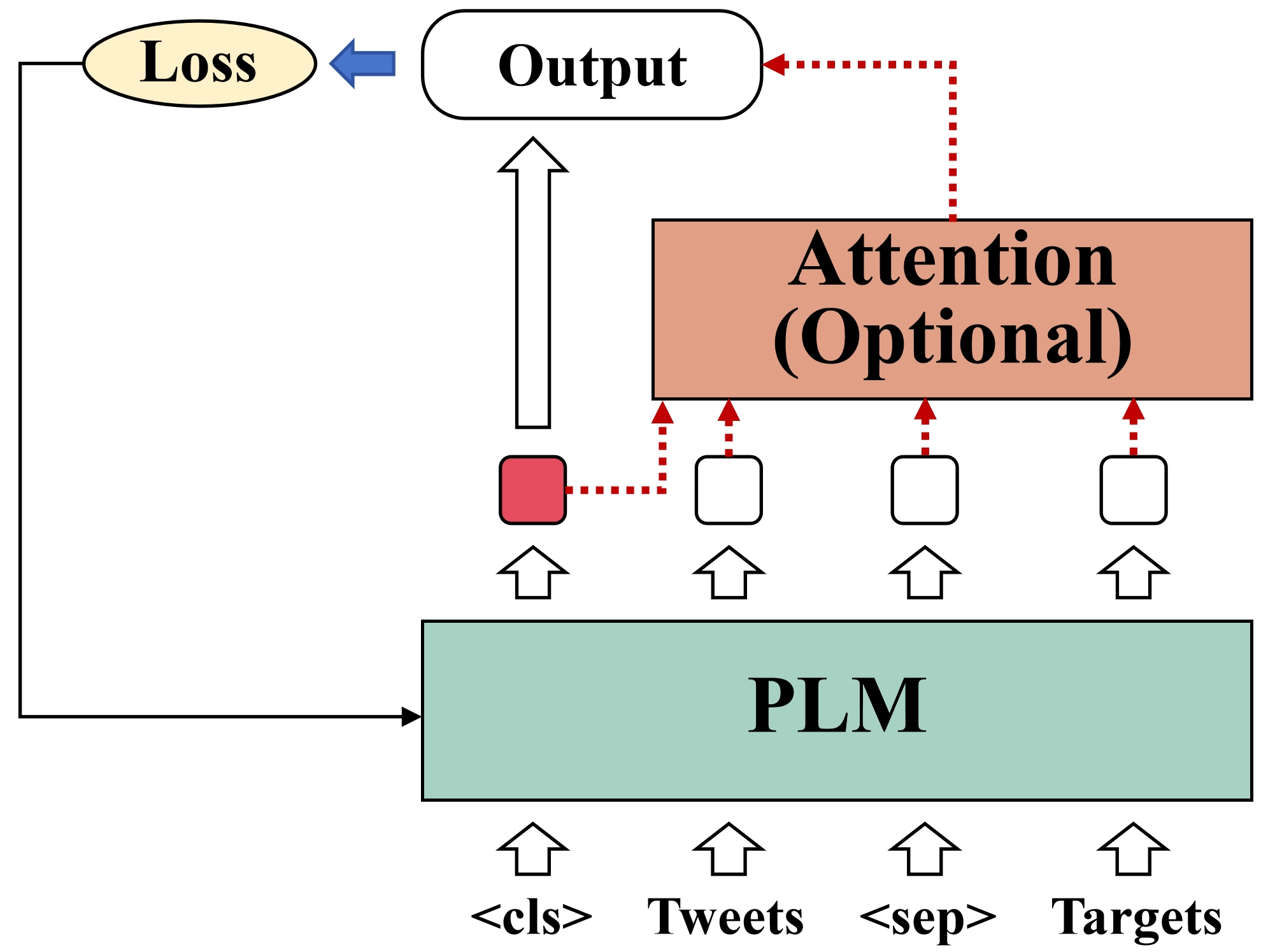}
\caption{Framework of fine-tuning approaches.} 
\label{PLMmethods}
\end{figure}

\subsection{Fine-Tuning Stance Detection Methods}
The advancement of deep learning techniques has led to significant performance improvements in text classification tasks using pre-trained language models (PLMs). As a result, PLMs have gained considerable attention in stance detection research. One effective approach is fine-tuning, which involves reformatting the input tweet and target into a new format, such as ``$<$CLS$>$ Tweet $<$SEP$>$ Target'', and then feeding it into a PLM to obtain the hidden state vector at the CLS position. This is followed by designing a loss function and training the model using labeled samples through backpropagation.  
The basic structure is shown in Fig.~\ref{PLMmethods}.

To enhance the applicability of PLMs in stance detection, some methods draw inspiration from the pre-training process of PLMs and retrain the PLM on additional social media data. For example, \citet{he2022infusing} retrained a PLM on social media tweet data.
Another approach is to use frozen PLM parameters, which is a common method due to its efficiency and minimal computational overhead. By adjusting a small portion of the model parameters and keeping the remaining parameters unchanged, PLMs can adapt to stance detection tasks while minimizing computational costs and data requirements ~\cite{li2024mitigating,gül2024stance,lan2023stance}. This method typically involves adding additional neural network layers on top of the CLS hidden state vector and only learning the additional parameters during training.

Some studies have also combined PLMs with traditional neural networks, such as RNNs and CNNs, to leverage the strengths of both approaches~\cite{li2022improved,niu2024challenge}. Furthermore, incorporating attention mechanisms with fine-tuning methods has been shown to significantly improve stance detection performance, leading to a large body of work focused on customizing attention mechanisms to address specific problems. A list of specific methods is provided in Table~\ref{PLMM}.

\begin{figure}
\centering
\includegraphics[width=0.65\linewidth]{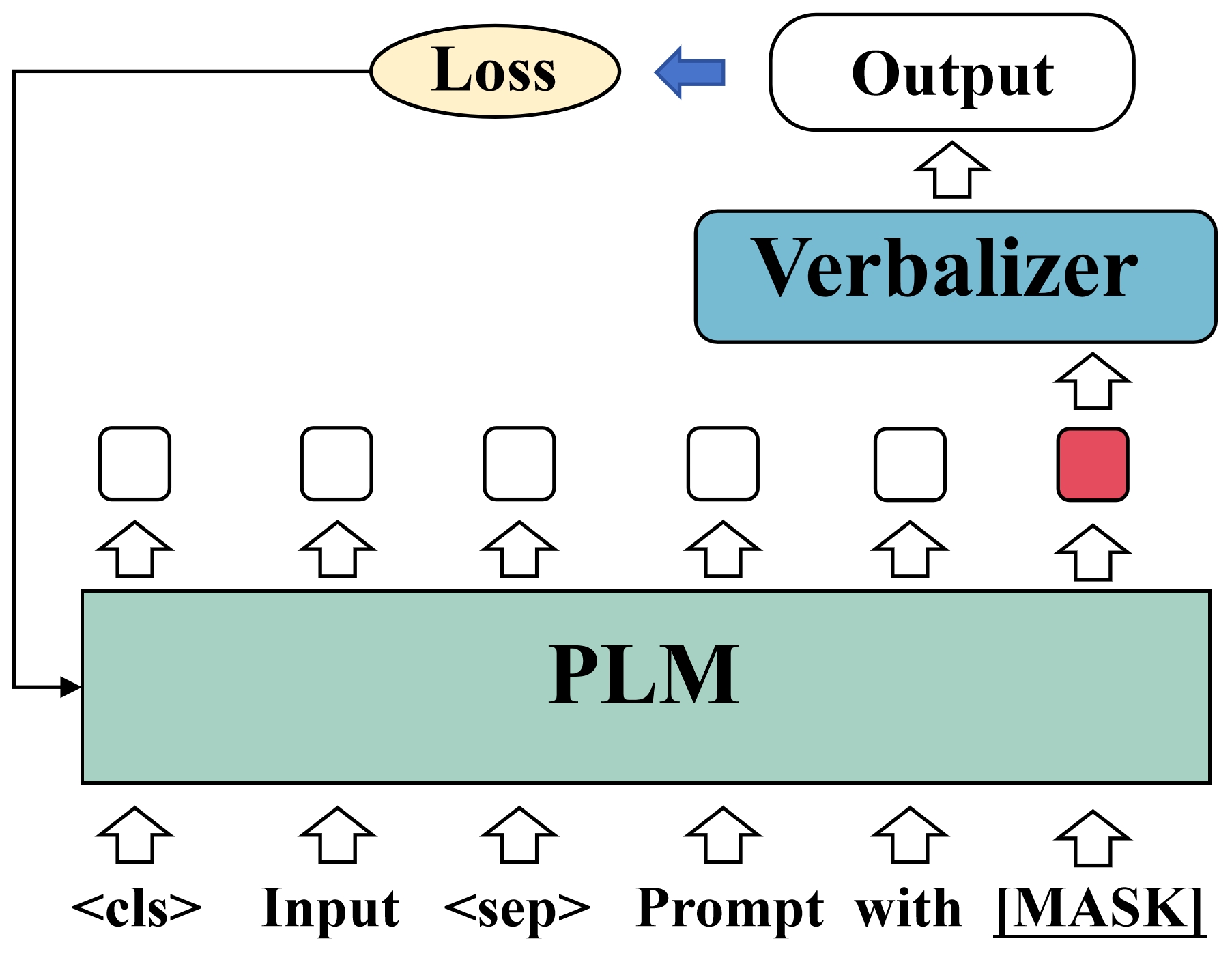}
\caption{Framework of prompt-tuning approaches.} 
\label{Ptuningmethods}
\end{figure}

\subsection{Prompt-Tuning Stance Detection Methods}
Prompt-tuning is an improvement over fine-tuning methods for stance detection. This approach formulates the stance detection task as a masked language modeling task. Specifically, prompt-tuning combines the given post and target with a template, typically written as ``The attitude to the $<$Target$>$ is \underline{[MASK]}''. The template is then packed with the input, and then fed into a PLM.
Subsequently, the hidden state at the mask position generates a prediction probability. This probability corresponds to a word or phrase defined in a verbalizer.
The basic structure is shown in Fig.~\ref{Ptuningmethods}.
To enhance performance, \citet{huang2023knowledge} proposed a method for automatically constructing a verbalizer tailored to the stance detection task.
\citet{wang-pan-2024-target-adaptive} is a flexible prompt-tuning approach that can be applied to various domains, thanks to its target-adaptive nature.
To address the challenge of multi-modal stance detection, \citet{liang2024multi} further introduced a visual prompt model.
A list of specific methods is provided in Table~\ref{PLMM}.

% \begin{figure*}[H] 
% 	\centering  %图片全局居中
% 	\vspace{-0.35cm} %设置与上面正文的距离
% 	\subfigtopskip=2pt %设置子图与上面正文或别的内容的距离
% 	\subfigbottomskip=2pt %设置第二行子图与第一行子图的距离，即下面的头与上面的脚的距离
% 	\subfigcapskip=-5pt %设置子图与子标题之间的距离
% 	\subfigure[original]{
% 		\label{level.sub.1}
% 		\includegraphics[width=0.32\linewidth]{FIG/zerocot.png}}
% 	\quad %默认情况下两个子图之间空的较少，使用这个命令加大宽度
% 	\subfigure[level=9]{
% 		\label{level.sub.2}
% 		\includegraphics[width=0.32\linewidth]{FIG/COT.png}}
% 	  %这里是空了一行，能够实现强制将四张图分成两行两列显示，而不是放不下图了再换行，使用\\也行。
% 	\subfigure[level=8]{
% 		\label{level.sub.3}
% 		\includegraphics[width=0.32\linewidth]{FIG/COLA.png}}
% 	\quad
% 	\subfigure[level=7]{
% 		\label{level.sub.4}
% 		\includegraphics[width=0.32\linewidth]{FIG/kasd.png}}
% 	\caption{ssss}
% 	\label{level}
% \end{figure*}

\section{Approaches based on Large Language Models}

Recently, the release of GPT-3.5 in November 30, 2022 has sparked significant interest in LLMs, such as the GPT-series, which have demonstrated impressive zero-shot performance on commonsense reasoning tasks \cite{binz2023using}. This has led to a new research direction in stance detection, focusing on utilizing LLMs. The way LLMs are employed has a substantial impact on the construction of stance detection models. A list of specific methods is provided in Table~\ref{LLMmethods}. 
% Therefore, this section will provide a detailed discussion on the two primary approaches to leveraging LLMs: direct prompting methods and large model augmentation methods.

\begin{figure}
\centering
\includegraphics[width=0.85\linewidth]{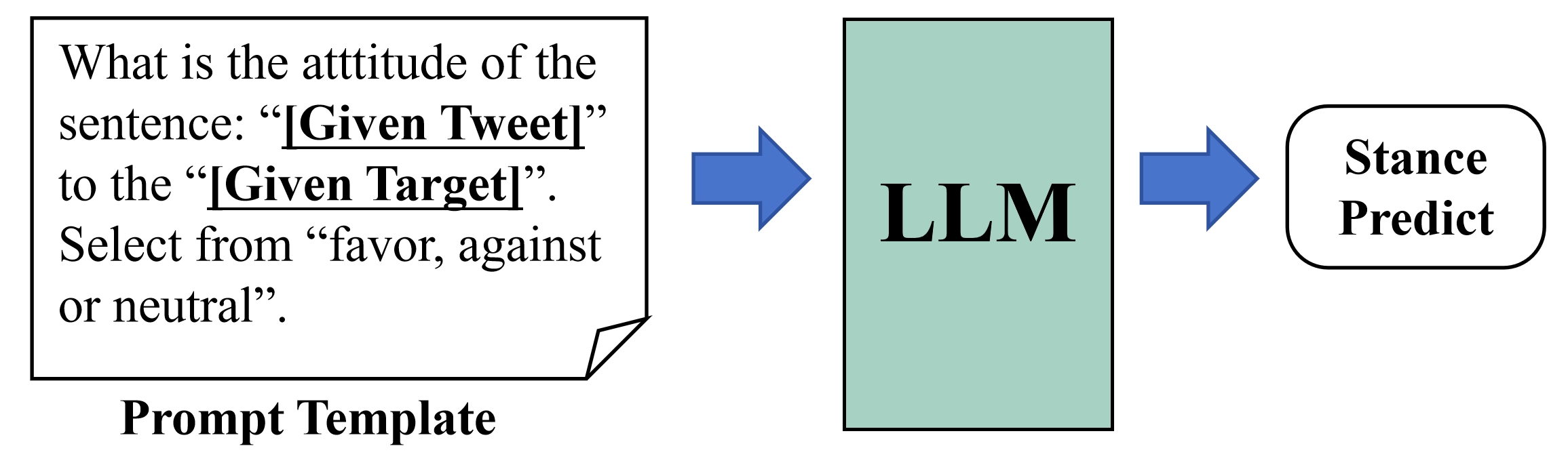}
\caption{Framework of zero-shot prompt methods.} 
\label{zsprompt}
\end{figure}

\begin{figure}
\centering
\includegraphics[width=1\linewidth]{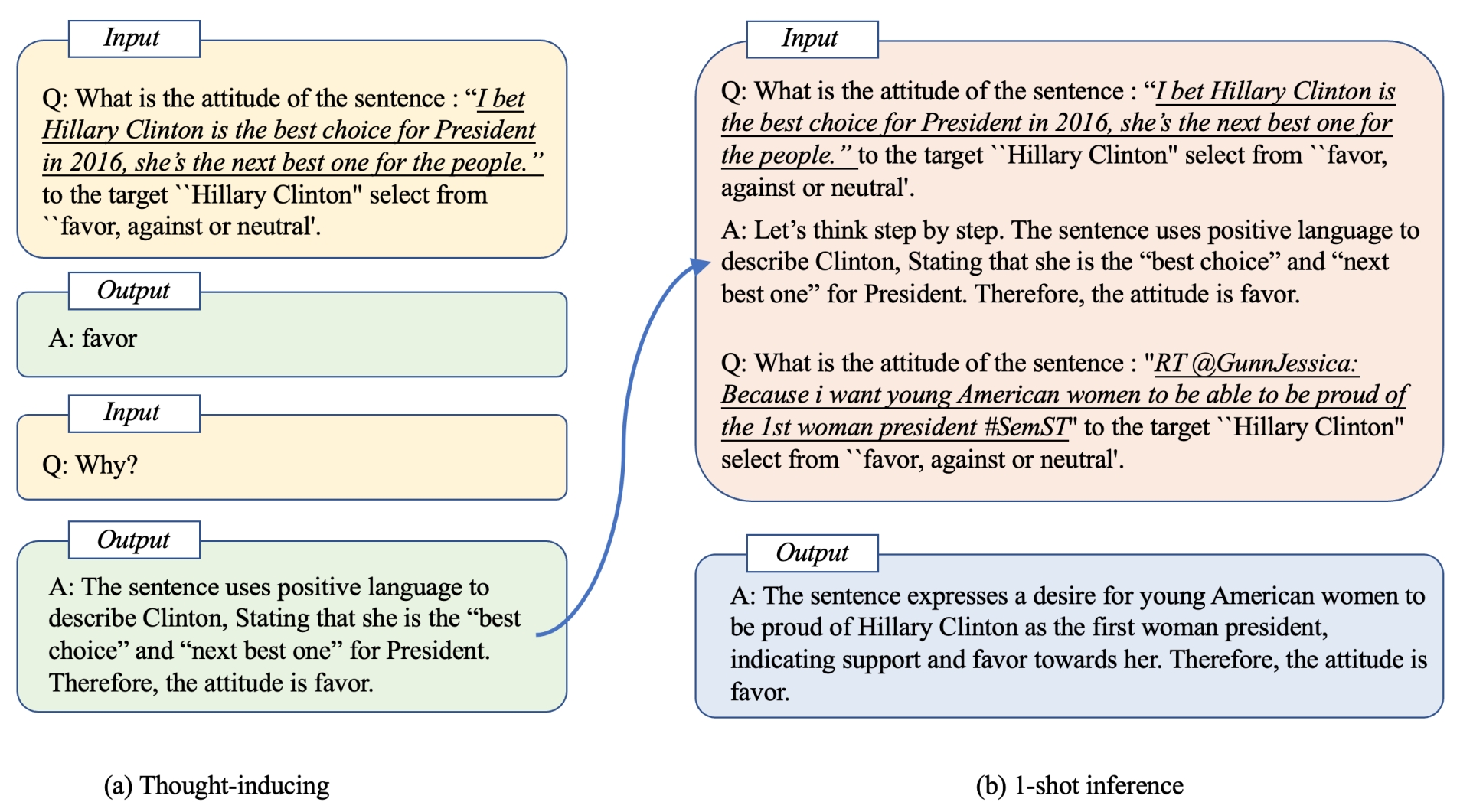}
\caption{Framework of chain-of-thought prompt method from \cite{zhang2023investigating}.} 
\label{cotprompt}
\end{figure}

\begin{figure}
\centering
\includegraphics[width=1\linewidth]{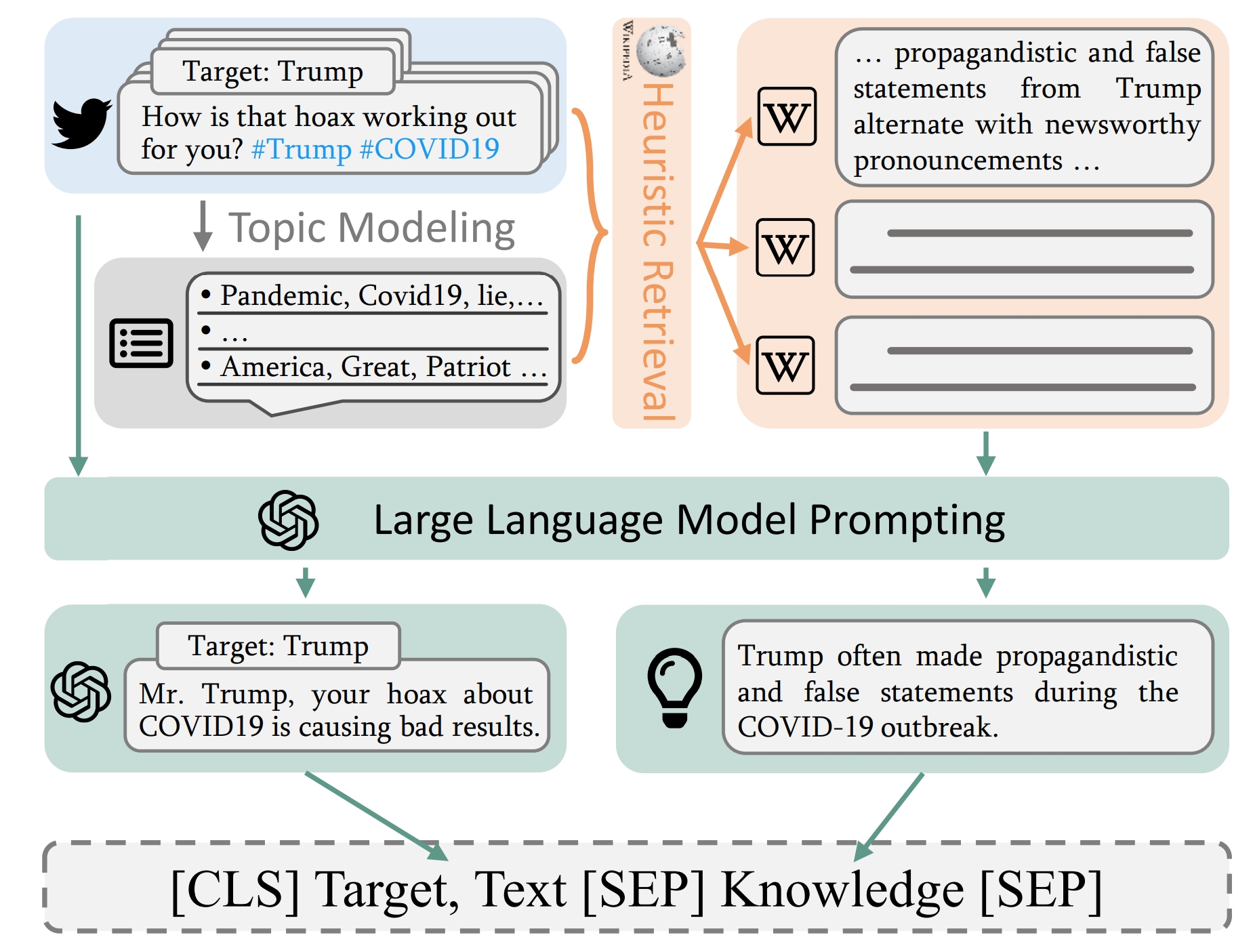}
\caption{Framework of background knowledge integrated prompt method from \cite{li2023stance}.} 
\label{bkprompt}
\end{figure}

% \subsection{Prompt-based Reasoning with LLMs}
\subsection{LLM-as-Predictors}

\textit{Utilizing their extensive internal knowledge and emerging reasoning abilities for stance detection.}

LLM-as-Predictors methods aim to leverage the inherent reasoning capabilities of LLMs by providing direct prompts, thereby treating LLMs as efficient knowledge experts ~\cite{zhang2022would,li2023stance,lan2023stance}. This approach enables the prediction of user stance without requiring training samples or further model training, making model deployment more efficient.

\textbf{Zero-shot prompt method}: \citet{zhang2022would} first proposed a zero-shot prompting method for stance detection, which is a simple yet effective approach. By crafting inquiry instructions, the LLM is directly asked to predict the stance polarity of the input. Experimental results demonstrate that this approach can achieve satisfactory stance detection accuracy. Fig.~\ref{zsprompt} illustrates the framework architecture.

\textbf{Chain-of-Thought prompt method}: Building on the zero-shot prompting method, \citet{zhang2023investigating} proposed a one-shot stance detection method based on thought chains. Fig.~\ref{cotprompt} shows the framework. By providing a pre-defined example for the LLM to learn from, the model is then asked to predict the stance of the input through inquiry. Experiments show that providing a single example significantly improves the model's prediction accuracy. This method also provides a potential prompting approach for future work.

% \textbf{Multi-agent method}: 
% The COLA \cite{lan2023stance} model views the LLM as an agent and enhances stance detection accuracy by allowing multiple independent LLMs to interact through prompting. 
% Building upon COLA, the DEEM method\cite{wang2024deem} further extends this approach by leveraging generated expert representations to facilitate semi-parametric reasoning in LLMs, thereby increasing the generalizability and reliability of the experts. Fig.~\ref{map} shows the framework.
% In this kind of approach, LLMs are assigned three roles: linguistic expert, domain specialist, and social media veteran. Each role analyzes the text from a distinct perspective, focusing on syntax and diction, characters and events, and platform-specific terminology, respectively. The combined insights from these roles help identify stance indicators in the text.
% Next, advocates for each potential stance category present arguments based on evidence from the preceding phase, forcing the LLMs to discern the latent logic connecting textual features and actual stances. The stance conclusion stage then determines the text's stance, integrating insights from both the text itself and the debates.

\textbf{Background knowledge integrated prompt method}: 
\citet{li2023stance} divides the prompting process into two stages, first acquiring the LLM's knowledge of the background through prompting, and then feeding the background knowledge and input into a second-stage LLM to predict stance polarity (Fig. \ref{bkprompt}).
\citet{weinzierl-harabagiu-2024-tree} further extends the prompting method to the multi-modal domain, demonstrating the effectiveness of prompting learning methods.

% \subsection{LLM-enhanced Stance Detection Methods}
\subsection{LLM-as-Enhancers}

\textit{Augmenting data and existing approaches with enhanced external knowledge and analytical prowess.}

% ====
Several studies argue that even LLMs often lack sufficient domain-specific knowledge to complete tasks, and the world knowledge embedded in LLMs can become outdated over time. Therefore, some works explore using sample prompts to elicit stance detection-related knowledge from LLMs, and then combine it with traditional deep learning methods, fine-tuning methods, or prompt-tuning methods to improve stance detection accuracy. We refer to this approach as LLM-enhanced fine-tuning methods ~\cite{zhang2024knowledge,ding2024cross1,zhang2024llm}.
This type of method typically involves at least two stages: first, it uses prompt learning to acquire relevant background knowledge; then, it inputs the acquired knowledge along with the original text into a trainable stance detection model.
This approach effectively combines the reasoning capabilities of LLMs with the learning capabilities of annotated samples, enhancing the model's zero-shot learning ability.

In the work of \citet{li2023stance}, in addition to directly prompting the model to acquire background knowledge and subsequently feeding it into the LLM to obtain the final stance (as mentioned above), the authors also proposed a fine-tuning approach in the second stage. Specifically, the extracted background knowledge is combined with the input tweet and target, and then fed into the fine-tuning method, which leverages labeled samples to learn through backpropagation.

\begin{figure}
\centering
\includegraphics[width=0.8\linewidth]{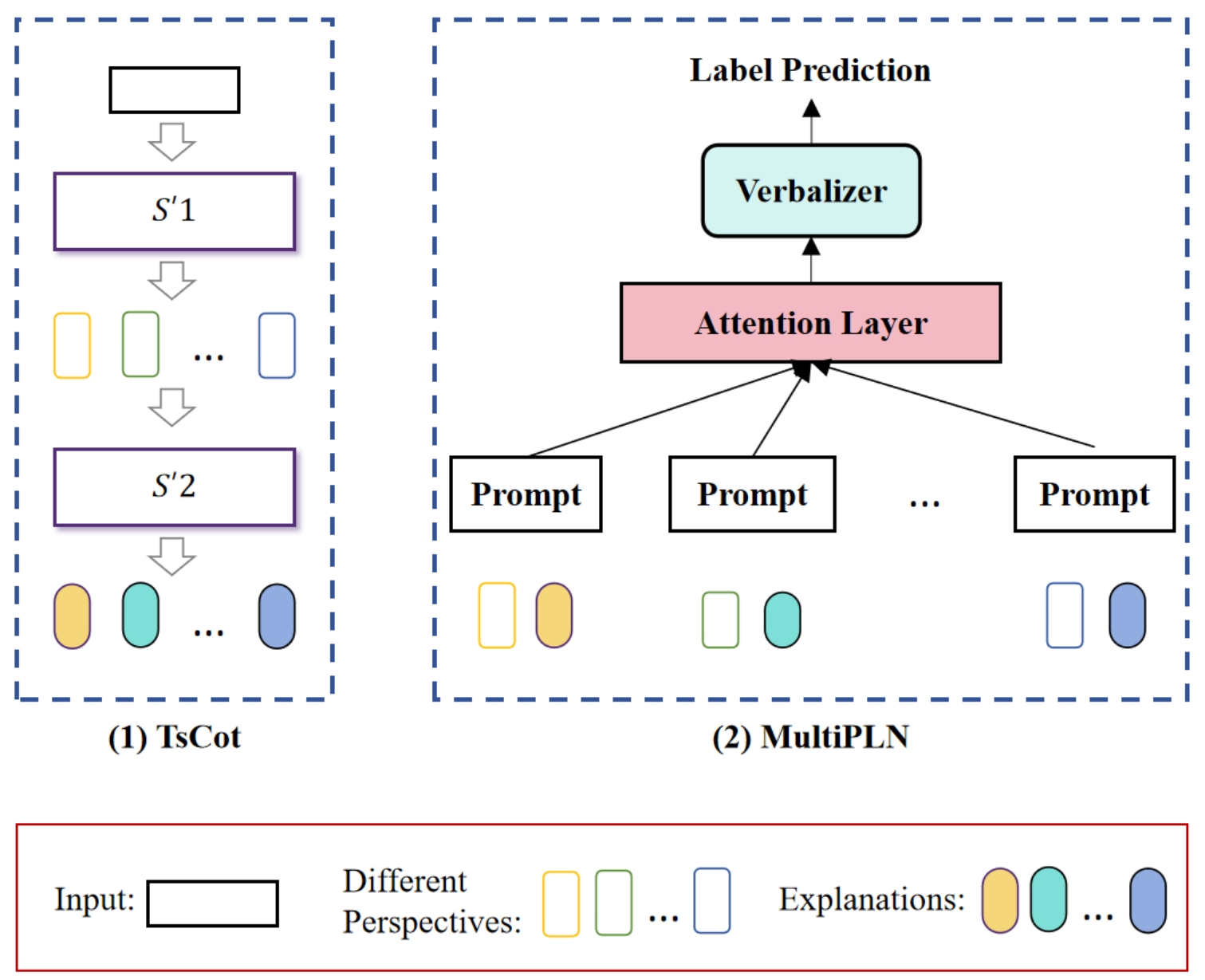}
\caption{Framework of MPPT \cite{ding2024cross1}.} 
\label{mpptf}
\end{figure}

% \textbf{Target-related knowledge enhanced fine-tuning method}: 
The MPPT method \cite{ding2024cross1} further expands the utilization of LLM knowledge by leveraging the CoT approach to extract the LLM's analysis of different perspectives on a target. 
Fig.~\ref{mpptf} shows the framework.
For example, for the target ``Trump'', the analysis perspectives might include ``politician, president, and Republican party member, etc.''. These perspectives can be regarded as domain-invariant knowledge that can be transferred to other targets, such as ``Hillary''. By incorporating the perspective information into the prompt-based fine-tuning method, MPPT enhances the accuracy and generalizability of stance detection across multiple targets.

\begin{figure}
\centering
\includegraphics[width=1\linewidth]{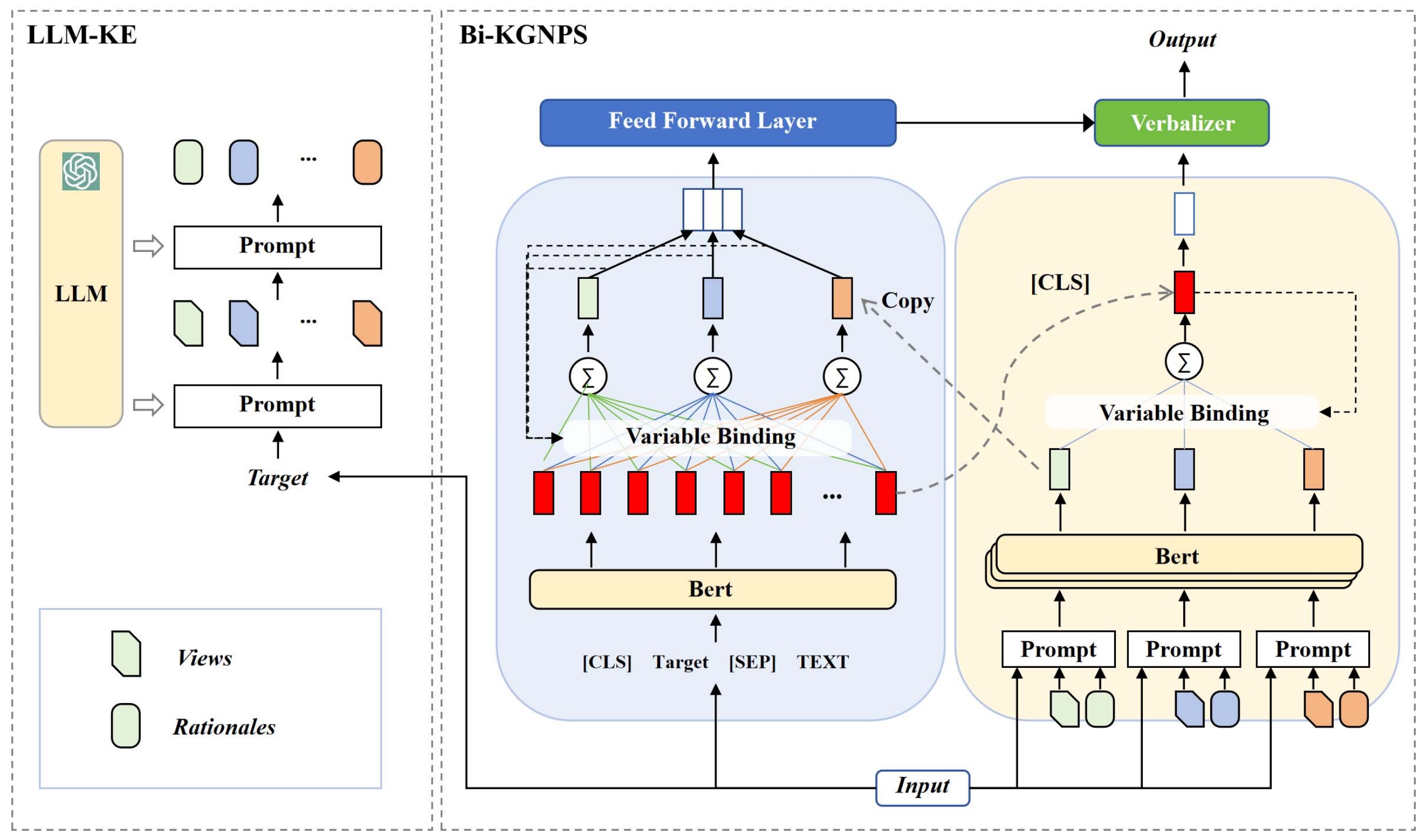}
\caption{Framework of KAI \cite{zhang2024knowledge}.} 
\label{kaikai}
\end{figure}

KAI~\cite{zhang2024knowledge} further extends the idea of leveraging LLMs to enhance reasoning by explicitly representing the logic of stance prediction using first-order logic rules. It is then combined with fine-tuning methods to improve the interpretability and accuracy of stance detection. Fig.~\ref{kaikai} shows the framework.
EDDA~\cite{ding2024edda} expands the generality of stance detection tasks by proposing a data augmentation-based approach. Using first-order logic rules as the foundation for data augmentation, EDDA generates new tweet-category sample pairs through prompt-based learning. Furthermore, by incorporating first-order logic rules as external knowledge, EDDA significantly improves the accuracy of current stance detection methods when combined with traditional fine-tuning models.

\begin{figure}
\centering
\includegraphics[width=0.8\linewidth]{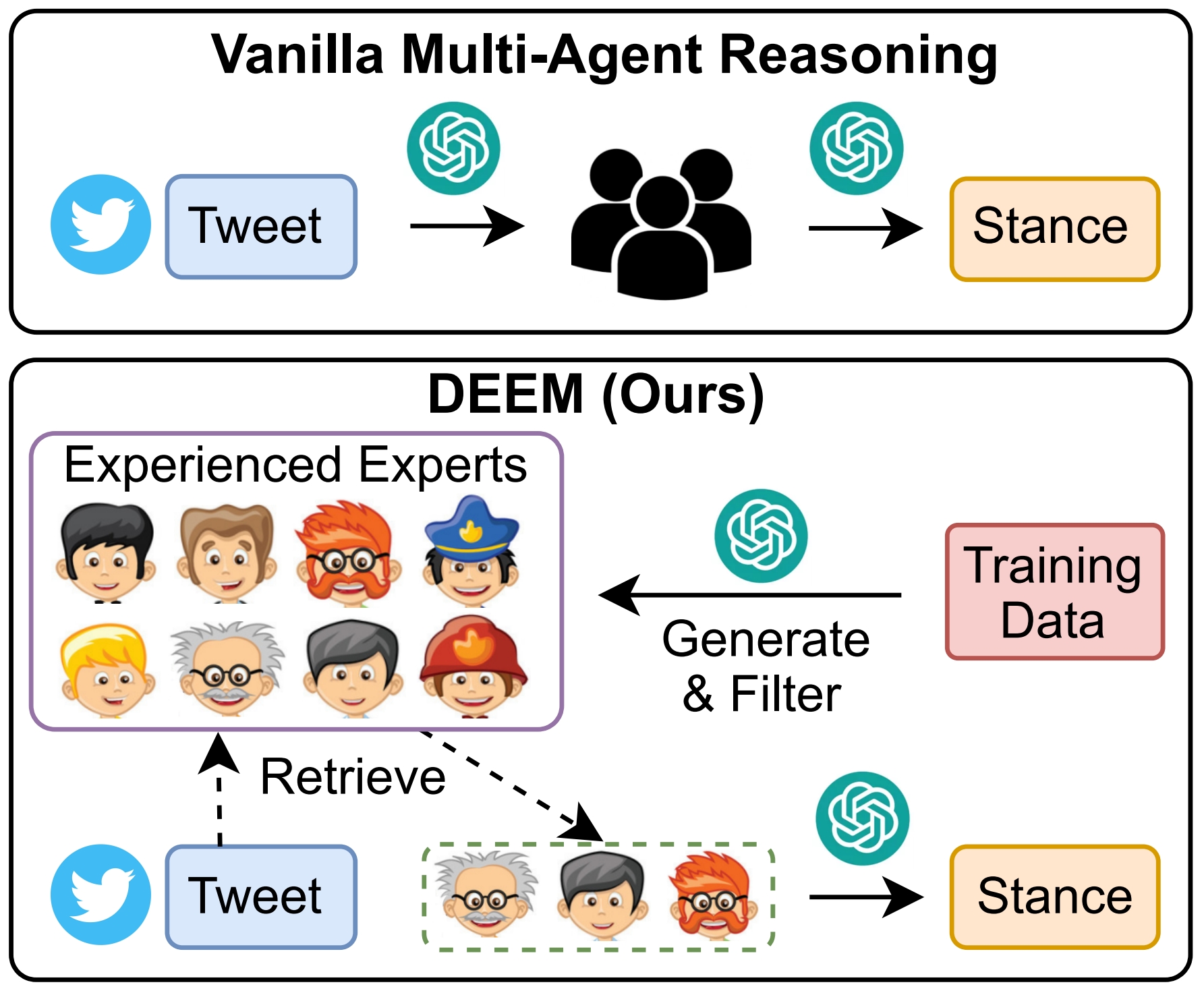}
\caption{Framework of multi-agent prompt method from \cite{wang2024deem}.} 
\label{map}
\end{figure}

\subsection{LLM-as-Agents}
\textit{Transcending conventional roles to actively engage in and transform stance detection.}

The COLA \cite{lan2023stance} model views the LLM as an agent and enhances stance detection accuracy by allowing multiple independent LLMs to interact through prompting. 
Building upon COLA, the DEEM method\cite{wang2024deem} further extends this approach by leveraging generated expert representations to facilitate semi-parametric reasoning in LLMs, thereby increasing the generalizability and reliability of the experts. Fig.~\ref{map} shows the framework.
In this kind of approach, LLMs are assigned three roles: linguistic expert, domain specialist, and social media veteran. Each role analyzes the text from a distinct perspective, focusing on syntax and diction, characters and events, and platform-specific terminology, respectively. The combined insights from these roles help identify stance indicators in the text.
Next, advocates for each potential stance category present arguments based on evidence from the preceding phase, forcing the LLMs to discern the latent logic connecting textual features and actual stances. The stance conclusion stage then determines the text's stance, integrating insights from both the text itself and the debates.

\section{Discussions and Future Work}

Building on previous work, we observe that the rapid advancement of LLMs has significantly improved the accuracy of traditional methods, achieving F1 scores of 80\% or higher on various tasks. This prompts us to predict that future research will move beyond current single-text or multi-modal content-level stance detection methods, expanding into more complex and realistic application scenarios. From the perspective of LLMs, we re-examine the development trajectory of stance detection tasks and propose several potential avenues for future research.

\subsection{Specializing in Social Media-Specific Data for Stance Detection}
Future research in stance detection should prioritize the development of datasets that more accurately reflect the complexities of real-world social media content. Social media platforms feature diverse and evolving content, posing a challenge for researchers to create datasets that accurately capture these characteristics. Although existing multi-modal and multi-turn dialogue stance detection datasets have partially simulated social media interactions, they are limited in number and scope. This highlights the need for future research to develop more comprehensive datasets that cover a broader range of social media formats, including comments, community interactions, and user-generated multimedia content such as images, videos, and text.
The development of such datasets faces several challenges. Firstly, the high heterogeneity and dynamic nature of social media data make it difficult to extract useful information. Users' expression styles are diverse and constantly evolving with social media trends. Secondly, annotating such data requires a significant amount of precise and labor-intensive work, and current automated annotation tools are insufficient for handling such complex data. Therefore, researching and developing efficient data processing and annotation techniques that can adapt to these new forms of data is essential for future research.

\subsection{Real-Time Stance Detection}

Currently, most stance detection tasks focus on static data from social media. However, in real-time stance detection, the primary goal is to immediately capture public opinion during live events such as political debates or breaking news. This requires continuously updating models to reflect current trends and sentiment shifts, enabling organizations to make timely decisions based on public reactions.

However, achieving real-time analysis poses several challenges. A major difficulty is handling the vast amounts of data streaming from social media and other platforms. Algorithms must be both fast and efficient, processing large volumes of data without compromising accuracy. Additionally, filtering out irrelevant or noisy data is crucial to ensure that analysis remains reliable and meaningful.

Another challenge is ensuring that models adapt to the rapid evolution of public discourse. As opinions change, models must remain accurate and relevant, which requires continuous updates and improvements. Scalability is also a concern, as systems need to effectively handle varying data loads and adapt to different languages or regions. Addressing these challenges is essential for effective real-time stance detection.

\subsection{User-Level Stance Detection}
Existing research on stance detection primarily focuses on content-level analysis, but user-level stance detection is crucial for understanding individuals' sustained attitudes towards various topics. To our knowledge, only two studies have explored user-level stance detection, both using non-human-annotated datasets, revealing the potential value and research gap in this area.
The goal of user-level stance detection is to capture and analyze users' overall attitudes across multiple interactions, rather than just a single text. This is particularly important for brand management, political analysis, and social movement research, as it can reveal changes in user attitudes and underlying trends. By understanding users' overall stances, we can more accurately predict their future behavior and decision-making.
However, achieving this goal is challenging. Data collection and privacy protection are significant hurdles, as user data is often scattered across multiple platforms and subject to strict privacy policies. Additionally, users' expressed stances may be inconsistent due to contextual changes, requiring complex models that can capture these changes. Furthermore, annotating user-level data is costly and time-consuming, necessitating the development of automated annotation techniques to improve efficiency.
In summary, user-level stance detection has immense application potential but faces numerous challenges. Future research must overcome these challenges in data acquisition, model design, and annotation efficiency to advance this field.

\subsection{Ethical Considerations and Bias Mitigation}
In stance detection, ethical considerations and bias mitigation are crucial, particularly when using LLMs. These models can amplify biases present in the training data, which may contain imbalanced and biased information. Developing debiased datasets and model training methods is essential but challenging.
The hallucination problem in LLMs, where they generate inaccurate or false information, is a key challenge. This can lead to incorrect stance detection results, affecting decision-making. Ensuring the reliability and accuracy of large model outputs is the first step in mitigating this issue. Researchers need to incorporate mechanisms into model training to detect and correct these hallucinations.
Debiasing methods in LLMs require special attention. While developing debiased datasets is one direction, it is difficult to achieve complete debiasing due to the massive scale of LLMs. Therefore, research should focus on the model training process, adopting dynamic adjustment and debiasing algorithms that can identify and adjust biased outputs in real-time, providing more fair results.
Moreover, transparency and explainability are crucial in LLMs. Due to their complexity, understanding their decision-making processes is a challenge. Developing tools and methods to explain model decisions can help users trust and accept model outputs.
In summary, as LLMs become increasingly applied in stance detection, the urgency to address bias and hallucination issues grows. Future research needs to innovate in model training, data processing, and result interpretation to ensure the ethicality, fairness, and reliability of stance detection technology.

\subsection{Cross-Platform Stance Detection}
Cross-platform stance detection is crucial for understanding users' overall attitudes and social dynamics. By integrating data from multiple social media platforms, this approach provides a more comprehensive analysis of user opinions. It can reveal cross-platform consistency and changes in stance, offering richer contextual information for decision-making.
However, this field faces numerous challenges. Firstly, standardizing and integrating data from different platforms is a complex task due to varying data formats and features. Additionally, users' stance expressions may differ across platforms due to platform-specific cultures and characteristics, requiring models to recognize and adapt to these differences.
Another significant challenge is data privacy and security. Cross-platform data collection may involve user privacy issues, necessitating secure and compliant data usage in accordance with laws and regulations. Furthermore, platform competition and data sharing restrictions can hinder data acquisition.
Technically, developing models that can handle multi-source heterogeneous data is a major challenge. This requires innovative algorithms and architectures to process text, images, and other multi-modal data simultaneously. Moreover, cross-platform stance detection models must be highly scalable to accommodate diverse platforms and languages.
In summary, cross-platform stance detection offers a new perspective on user attitude research but demands breakthroughs in data integration, privacy protection, and technical development. Future research must innovate in these areas to drive the development and application of this field.

\subsection{Virtual-Real Fusion Stance Detection}
The demand for virtual-real fusion stance detection arises from the pursuit of comprehensively understanding user attitudes. In today's interconnected world, users' opinions and behaviors are influenced by both online interactions and real-world contexts. By combining virtual and real-world data, we can gain deeper insights into user behavior.
Virtual-real fusion in stance detection provides a more comprehensive and realistic user portrait. By integrating virtual and real-world data, we can obtain more accurate user attitude analysis. This approach can reveal users' behavioral consistency and differences across various contexts, offering a new perspective on complex social dynamics.
Virtual-real fusion significantly improves tasks, particularly in social behavior research, personalized services, and smart city management. For instance, researchers can analyze the relationship between online mobilization and offline participation in social movements, enabling a better understanding of social change drivers. By combining online preferences and offline activity data, personalized services can provide more dynamic and tailored recommendations, enhancing user experience.
Moreover, virtual-real fusion enables more personalized and dynamic applications. By integrating online and offline information in real-time, systems can adapt to users' changing needs and environments, providing more accurate services and suggestions.
However, this field faces several challenges. Firstly, data integration complexity is a major obstacle. Online and offline data differ significantly in format and content, requiring effective integration methods. Additionally, collecting and using offline data while respecting user privacy is an important ethical concern.
Technically, achieving virtual-real fusion stance detection requires innovative algorithms that can handle and analyze multi-source heterogeneous data, including text, images, and sensor data. Models must also possess dynamic updating capabilities to adapt to rapidly changing real-world and virtual environments.
Finally, virtual-real fusion stance detection has vast application potential, including smart cities and augmented reality experiences. Therefore, research must delve deeper into application-level exploration to fully exploit this method's potential. By addressing these challenges, virtual-real fusion stance detection can provide new insights into understanding and predicting user behavior.

\section{conclusion}
This comprehensive survey has explored the rapidly evolving field of stance detection on social media, emphasizing the integration of large language models (LLMs) and their prospective contributions to understanding public sentiment on complex topics. First, the survey  presented a variety of targets for stance detection, including entities, events, and claims, reflecting the multifaceted nature of social media discussions.
Second, this paper reviewed both traditional and novel methodologies in stance detection, highlighting the transition from feature-based machine learning techniques to more advanced deep learning and LLM-based approaches. In particular, the utility of LLMs like the GPT series in stance detection was critically analyzed, acknowledging their potential to revolutionize the field by leveraging their deep contextual understanding capabilities.
Moreover, the paper addressed emerging avenues such as the detection of unexpressed stances and the application of stance detection tools in new domains like fake news identification. These applications underscore the broader relevance and utility of stance detection beyond academic research.
In conclusion, the survey underscored the need for ongoing research to develop more generalized and robust stance detection models capable of handling the dynamic and diverse nature of social media. It called for the creation of new, more comprehensive datasets and the continued exploration of NLP and social computing techniques. These efforts are crucial for advancing the field and for better harnessing the power of stance detection to inform decision-making processes in various sectors.

\ifCLASSOPTIONcaptionsoff
\newpage
\fi

\bibliographystyle{IEEEtranN}
% \bibliographystyle{IEEEtran}
% \bibliography{ref}
\bibliography{modified_file}
% \bibliographystyle{IEEEtran}
% \bibliography{IEEEabrv,ref}

\end{document}